\PassOptionsToPackage{table}{xcolor}

\documentclass[acmsmall,nonacm]{acmart}
\AtBeginDocument{%
  \setlength{\abovedisplayskip}{6pt}
  \setlength{\belowdisplayskip}{6pt}
  \setlength{\abovedisplayshortskip}{4pt}
  \setlength{\belowdisplayshortskip}{4pt}
}

\renewcommand\footnotetextcopyrightpermission[1]{}
\setcopyright{none}
\usepackage{amsmath}
\usepackage{amsthm}
\usepackage{mathrsfs}
\usepackage{booktabs}
\usepackage{tabularx}
\usepackage{makecell}
\usepackage{multirow}
\usepackage{threeparttable}
\usepackage{longtable}
\usepackage{array}
\usepackage{lscape}

\usepackage{enumitem}
\usepackage{textcomp}
\usepackage{soul}
\usepackage[normalem]{ulem}
\usepackage{microtype}
\usepackage{manyfoot}

\usepackage{graphicx}
\usepackage{tikz}
\usetikzlibrary{positioning}
\usepackage{subcaption}

\usepackage{algorithm}
\usepackage{algpseudocode}
\usepackage{listings}

\usepackage[title]{appendix}

\newcommand{\botrule}{\bottomrule}

\hypersetup{
  colorlinks=true,
  linkcolor=blue,
  citecolor=blue,
  urlcolor=blue
}

\usepackage{etoolbox}
\makeatletter
\AtBeginDocument{%
  \patchcmd{\@typeset@author@line}{\par\noindent}{\par\centering}{}{}%

  \patchcmd{\@mktitle@i}{\raggedright}{\centering}{}{}%
  \patchcmd{\@mktitle@i}{\noindent\@titlefont}{\@titlefont\centering}{}{}%
}
\makeatother

\begin{document}

\title[Feature-Enhanced GNNs for Classification
of Synthetic Graph Generative Models]{Feature-Enhanced Graph Neural Networks for Classification of Synthetic Graph Generative Models: A Benchmarking Study}

\author{Janek Dyer}\email{2406731@uad.ac.uk}

\author{Jagdeep Ahluwalia}\email{j.ahluwalia@abertay.ac.uk}
\author{Javad Zarrin}\authornote{Corresponding author}\email{j.zarrin@abertay.ac.uk}

% \affiliation{%
%   \institution{Abertay University}
%   \city{Bell Street, DD1 1HG Dundee}
%   \country{Scotland, UK}
% }

\affiliation{%
  \institution{Abertay University}
  \streetaddress{Bell Street}
  \city{Dundee}
  \state{Scotland}
  \postcode{DD1 1HG}
  \country{UK}
}

\renewcommand{\shortauthors}{Dyer et al.}

\begin{abstract}
The ability to discriminate between generative graph models is critical to understanding complex structural patterns in both synthetic graphs and the real-world structures that they emulate. While Graph Neural Networks (GNNs) have seen increasing use to great effect in graph classification tasks, few studies explore their integration with interpretable graph theoretic features.
This paper investigates the classification of synthetic graph families using a hybrid approach that combines GNNs with engineered graph-theoretic features.
We generate a large and structurally diverse synthetic dataset comprising graphs from five representative generative families – Erdős–Rényi, Watts-Strogatz, Barabási-Albert, Holme-Kim, and Stochastic Block Model. These graphs range in size up to $1\times10^4$ nodes, containing up to $1.1\times 10^5$ edges. A comprehensive range of node and graph level features is extracted for each graph and pruned using a Random Forest based feature selection pipeline. The features are integrated into six GNN architectures: GCN, GAT, GATv2, GIN, GraphSAGE and GTN. Each architecture is optimised for hyperparameter selection using Optuna. Finally, models were compared against a baseline Support Vector Machine (SVM) trained solely on the handcrafted features.
Our evaluation demonstrates that GraphSAGE and GTN achieve the highest classification performance, with $98.5\%$ accuracy, and strong class separation evidenced by t-SNE and UMAP visualisations. GCN and GIN also performed well, while GAT-based models lagged due to limitations in their ability to capture global structures. The SVM baseline confirmed the importance of the message passing functionality for performance gains and meaningful class separation.
Our work contributes a scalable and reproducible benchmarking framework that highlights the value of integrating classical graph theory with GNN-based learning for large scale synthetic graph classification.
\end{abstract}

\keywords{Graph Neural Networks, Graph Classification, Synthetic Graphs, Graph Theory}

%%
%% This command processes the author and affiliation and title
%% information and builds the first part of the formatted document.
\maketitle

\section{Introduction}\label{intro}
From mapping social networks and optimising logistics, to designing novel pharmaceuticals and detecting financial fraud, graph-based representations are fundamental to modelling a wide array of complex physical and conceptual systems. By effectively representing systems through nodes and their connecting edges, these systems can be analysed and interpreted based on the inherent structural properties of their graphs. The ability to accurately classify these graphs holds significant real-world and scientific importance.

An early application of graph classification comes from Leonhard Euler’s approach to the Königsberg Bridge problem\cite{biggs1986graph}\cite{euler1736bridge}, whereby Euler showed that no walk through the city of Königsberg would allow for each of the seven bridges within to be crossed once and only once. Euler’s proof relied on reducing the map of the city to a graph, and the observation that each node was of odd degree. The result of the proof was that no solution existed to the problem. Graphs for which a solution to the problem exists are now known as Eulerian.

While Euler’s approach was groundbreaking, and pivotal in terms of introducing formal graph theory, many networks involved in current graph classification problems contain significantly more than seven edges, and as such do not lend themselves to manual consideration. Structures may also differ in subtle ways, requiring analysis of clustering patterns or eigenvectors to facilitate classification.
The non-Euclidean structure and arbitrary size of graphs present significant challenges in approaches to classification. Real-world networks can also contain significant noise , which obscures underlying structures, as well as imbalanced class distributions. In such cases, classification methods can benefit from evaluation upon inherently clean and balanced synthetic datasets\cite{tsitsulin2022synthetic}\cite{ju2024surveygraphneuralnetworks}.

Synthetic datasets have a significant advantage over real-world datasets in that they have arbitrary scale and complexity, as well as well-defined structures\cite{darabi2025synthetic}\cite{palowitch2022synthetic}. This allows for evaluation of classification approaches without concerns regarding data cleansing or volume of quality annotated data\cite{paulin2023synthetic}. The obvious disadvantage of using synthetic datasets is that the lack of noise compared to real-world datasets means that results obtained may not map directly between synthetic and organic data. However, the practicalities and clean conditions afforded by synthetic datasets make them a valuable tool for benchmarking and evaluation.

This study evaluates six diverse and widely used GNN architectures on a graph classification task. The architectures featured were selected based on their prevalence in current graph-learning literature, as well as their diversity in their individual message-passing mechanisms or attention strategies. Each GNN evaluated in this study represents a distinct approach to graph classification. The controlled nature of this study allows for discrete evaluation of these approaches in a stable experimental environment.

Five generative graph families are explored for classification. These were chosen due to their use as canonical models that capture distinct real-world structural traits, as well as providing potential for overlap in their structures. Considered in combination, the selected graphs represent a challenging classification task with overlapping topological properties that necessitates deep structural insight on the part of the classifier.

The aim of this study is to evaluate the discriminative power of different GNN architectures between graph families that exhibit nuanced and overlapping structural patterns. By integrating lightweight handcrafted features with expressive message passing mechanisms, the study seeks to evaluate the global classification performance of hybrid GNN models. Specifically, we examine whether architectural differences such as local vs global attention, and inductive vs transductive reasoning, allow for consistent differences in performance across structurally complex synthetic graphs. 

We generate a synthetic dataset of 2000 graphs, comprised of 400 examples from each of the five different generative families investigated. Recursive feature ablation is performed using Random Forest classification, with the most discriminative features being provided to the GNN models along with the adjacency matrix information. Hyperparameter optimisation is performed using Optuna\cite{akiba2019optuna} to establish strong candidate models for each of the investigated architectures. The best performing candidate model for each architecture is then evaluated across a full training cycle. Baseline performance is established through use of a Support Vector Machine model, allowing for comparison of the specialised GNN architectures to a generalised model for graph classification tasks.

In summary, this study makes the following key contributions. We perform an extensive empirical evaluation of state-of-the-art GNN architectures on a novel synthetic graph classification task, benchmarking them in a controlled environment to isolate architectural strengths. We introduce a scalable hybrid graph classification pipeline that integrates handcrafted graph-theoretic features with message-passing mechanisms, and assess its impact on classification accuracy and interpretability. Thus, this research provides a clear benchmark to help practitioners select the most suitable GNN architecture where graph structure is the key discriminating factor.

The remainder of this paper is structured as follows. Section \ref{rel_work} reviews relevant literature in graph classification approaches and GNN architecture design. Section \ref{graph_families} details the generative families used to create the dataset, providing formal definitions and providing an overview of their structural properties. Sections \ref{data_gen} and \ref{feat_select} outlines the data generation and feature engineering process. Section \ref{cand_models} describes the candidate models and architectural variants. Section \ref{training_optim} covers the hyperparameter optimisation and training protocols. Section \ref{eval} presents detailed evaluation results across metrics and visualisation approaches. Section \ref{disc} provides detailed discussion of key findings and limitations. Section \ref{conc} concludes with a summary and directions for future work.
 
\section{Related Work}\label{rel_work}
Graph classification is a well-established task, with approaches ranging from analysis of handcrafted features to cutting-edge deep learning methods. Throughout this section, we seek to highlight the gap in systematic, comparative benchmarking of diverse GNN architectures on structurally similar synthetic graphs that this study addresses by reviewing the current literature, with an emphasis on Graph Neural Networks (GNNs) and benchmarking efforts.

Manually engineered graph features can be defined at the local and global levels. Local features refer to attributes of individual nodes, such as degree, clustering coefficient, and eigenvector centrality. These features allow for insight into neighbourhoods and sub-structures of the graph, allowing for deep understanding of localities. Global features describe the graph as a whole. These include attributes such as shortest path length and degree variance, capturing information about the structure as a whole. Both global and local features may be used for classification purposes and provide a strong sense of interpretability. However, their use scales poorly in the context of larger networks representative of complex real-life systems.
Statistical approaches to network classification provided advances in the field, with methods such as the Weisfeiler-Lehman test\cite{weisfeiler1968wl} allowing for insight into isomorphism between graphs. The graph isomorphism problem involves determining whether two graphs are structurally identical, even if their node labels differ. Put simply, the problem asks if there exists a restructuring of node labels that results in identical graphs. Machine learning graph kernel methods would later integrate the Weisfeiler-Lehman isomorphism test, resulting in the Weisfeiler-Lehman subtree kernel\cite{shervas2011weis}. These approaches seek to make use of the expressiveness afforded by the Weisfeiler-Lehman test, while minimising computational overheads. Kernel methods for graph isomorphism have proven powerful but can scale poorly with graph size.  While conceptually simple, the lack of a known polynomial time solution to the isomorphism problem makes it a significant computational challenge for large or complex graphs. The practical result of this in terms of kernel methods for isomorphism is that a graph kernel with polynomial time complexity will be unable to accurately identify all pairs of isomorphic graphs\cite{borg2020kernel}. 

Feature selection remains crucial for graph kernel methods, although the focus is more on the identification of relevant subtrees of graphs for aiding classification, as opposed to traditional hand-crafted features\cite{tan2010featselect}.
Recursive Neural Networks were initially used for acyclic network classification\cite{sperduti1997supervise}, which marked a shift away from feature-based classification. Further studies have employed feed-forward neural networks to address cyclic structures\cite{Micheli2009neural}. While effective, these approaches still struggled to operate efficiently at scale. Advancements in deep neural networks resulted in the advent of Convolutional Neural Networks (CNNs), which offered the ability to learn about the locality of nodes, aggregating neighbourhood information without having to directly access the neighbouring nodes \cite{tixier2019graph}. However, CNNs fail to generalize for graphs since they assume a grid-like structure with fixed spatial relationships. While this assumption makes them well-suited for image-based tasks, they are less effective when applied to the non-Euclidean structures of graph networks.
Graph Neural Networks (GNNs) directly address this limitation and represent a paradigm shift in network analysis. GNNs can learn directly from the graph structures themselves, reducing reliance on computationally expensive features by iteratively aggregating and transforming information from each node’s neighbours and encoding both topology and attributes into their representations. At the core of this learning process is the method of Message Passing. Message Passing overcomes the challenges posed by graph-like structures by allowing nodes to exchange information within their localities and aggregate this information to update their own embeddings. By passing messages across the graph, both local and global structures can be identified, allowing for granular approaches to classification.
A generalised formulation of the message passing operation to obtain updated embeddings can be described as follows. For a node $i$ with current embedding $x_i$ and neighbourhood $\mathcal{N}(i)$, the updated embedding $x'_i$ can be written as:
\begin{equation}
 x'_i
   = \mathrm{UPDATE }\!\Bigl(
        x_i,\;
        \mathrm{AGGREGATE }\!\bigl(\{\,x_j: j \in \mathcal{N}(i)\}\bigr)
     \Bigr)   
\end{equation}
The AGGREGATE function generates a message $m_i$  based on neighbourhood node embeddings, that is passed to the UPDATE function to combine with $x_i$ to provide the updated embedding $x'_i$. 

\begin{figure}[H]
\centering
\begin{subfigure}[b]{0.475\textwidth}
    \includegraphics[width=\linewidth]{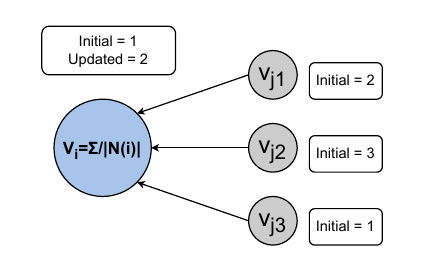}
    \caption{GCN - Static neighbourhood equal-weighting mean}
\end{subfigure}
\hfill
\begin{subfigure}[b]{0.475\textwidth}
    \includegraphics[width=\linewidth]{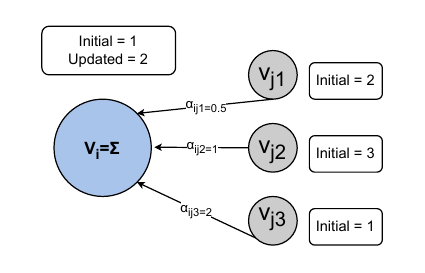}
    \caption{GAT - Static neighbourhood attention weighting}
\end{subfigure}

\begin{subfigure}[b]{0.475\textwidth}
    \includegraphics[width=\linewidth]{GAT_dia.pdf}
    \caption{GATv2 - Dynamic neighbourhood attention weighting}
\end{subfigure}
\hfill
\begin{subfigure}[b]{0.475\textwidth}
    \includegraphics[width=\linewidth]{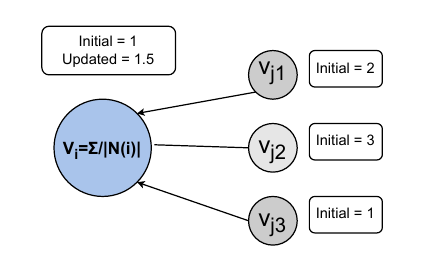}
    \caption{GraphSAGE - Sampled neighbourhood mean}
\end{subfigure}

\begin{subfigure}[b]{0.475\textwidth}
    \includegraphics[width=\linewidth]{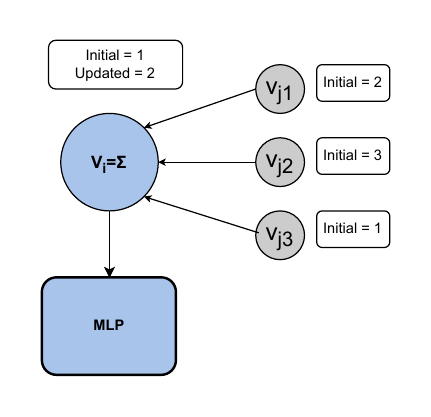}
    \caption{GIN - Sum aggregation and MLP}
\end{subfigure}
\hfill
\begin{subfigure}[b]{0.475\textwidth}
    \includegraphics[width=\linewidth]{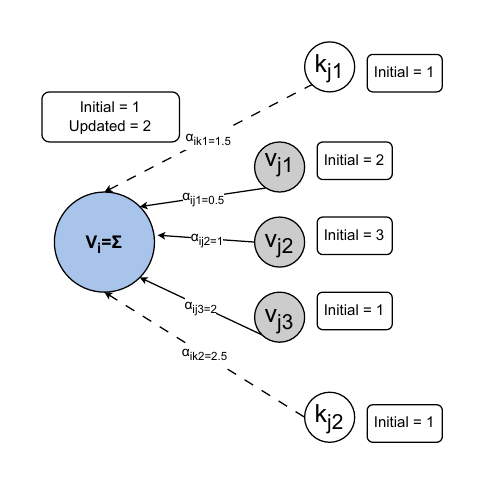}
    \caption{GTN - Global attention weighting}
\end{subfigure}
\caption{Visual summary of aggregation strategies for each featured GNN model.}
\label{fig:agg_strats}
\end{figure}

Graph Neural Networks comprise a diverse range of architectures, differing primarily in the formulations of their AGGREGATE and UPDATE functions. Figure \ref{fig:agg_strats} illustrates different aggregation mechanisms employed by the key architectures evaluated in this study. As illustrated, the featured architectures employ fundamentally different strategies to update their node embeddings. 
The Graph Convolutional Network (GCN)\cite{kipf2017gcn} uses simplified spectral convolution to assign an average weight to neighbours, whereas Graph Attention Networks (GAT)\cite{velivckovic2018gat} employ a static attention mechanism that allows for nodes to weight the importance of their neighbours. This mechanism is developed to allow for dynamic attention weightings in the GATv2 model\cite{brody2021gatv2}. Graph Sample and Aggregate (GraphSAGE)\cite{hamilton2017sage} models, meanwhile, aggregate feature information from a sample of neighbours, using inductive reasoning to generate embeddings for unseen nodes. Graph Isomorphism Networks (GIN)\cite{xu2019pgin}, seek to match the expressiveness of the Weisfeiler-Lehman test\cite{weisfeiler1968wl} through a combination of sum aggregation and updates using a multi-layer perceptron (MLP). Graph Transformer Networks (GTN)\cite{shi2021masked} seek to expand upon the attention mechanism functionality, extending the mechanism to cover all nodes in a graph. This allows for improved understanding of long range dependencies at the cost of increased computation.

\par
Given their distinct architectural strategies, the performance of each GNN variant is highly dependant upon the specific characteristics of the dataset it is applied to. Factors such as graph density, size, and the nature of its structural properties impact the performance capabilities of each architectural approach. This inherent task and dataset-specific nature of performance makes direct comparison challenging, and underscores the importance of evaluating models across a range of benchmarks.
Table \ref{tab:gnn_benchmarks} summarises reported accuracies achieved by each featured GNN architecture in node and graph classification tasks across a range of benchmarks:
\par
\begin{table}[htbp]
\caption{Performance of GNN architectures on graph and node classification benchmark datasets.}
\label{tab:gnn_benchmarks}
\begin{tabular*}{\textwidth}{@{\extracolsep{\fill}} l l l l l}
\toprule
\textbf{Architecture} & \textbf{Dataset} & \textbf{Task Type} & \textbf{Accuracy (\%)} & \textbf{Study} \\
\midrule
GCN         & IMDB-B\cite{yanardag2015bench}   & Graph & 73.60 & Job et al. (2024)\cite{job2024causalbench} \\
            & Reddit-B\cite{yanardag2015bench} & Graph & 91.25 &                   \\
\addlinespace
GAT         & IMDB-B\cite{yanardag2015bench}   & Graph & 72.00 &                   \\
            & Reddit-B\cite{yanardag2015bench} & Graph & 90.85 &                   \\
\addlinespace
GIN         & IMDB-B\cite{yanardag2015bench}   & Graph & 72.50 &                   \\
            & Reddit-B\cite{yanardag2015bench} & Graph & 88.75 &                   \\
\addlinespace
GraphSAGE   & IMDB-B\cite{yanardag2015bench}   & Graph & 73.00 &                   \\
            & Reddit-\cite{yanardag2015bench} & Graph & 77.20 &                   \\
\midrule
GAT         & ogbn-products\cite{hu2020ogb} & Node  & 79.04 & Brody et al. (2022)\cite{brody2021gatv2} \\
            & ogbn-arxiv\cite{hu2020ogb}    & Node  & 71.59 &                     \\
\addlinespace
GATv2       & ogbn-products\cite{hu2020ogb} & Node  & 80.63 &                     \\
            & ogbn-arxiv\cite{hu2020ogb}    & Node  & 71.78 &                     \\
\midrule
GTN         & ogbn-products\cite{hu2020ogb} & Node  & 82.56 & Shi et al. (2021)\cite{shi2021masked} \\
            & ogbn-arxiv\cite{hu2020ogb}    & Node  & 73.11 &                   \\
\bottomrule
\end{tabular*}
\end{table}
\par

As shown in Table \ref{tab:gnn_benchmarks}, performance differences in node and graph classification tasks between GNN variants can often reflect the structural biases of the specific datasets. This can make direct comparison difficult, as the noise inherent in real-world datasets limits insight into the true discriminative power of the architectures\cite{tsitsulin2022synthetic}. This motivates evaluation of GNN architectures on controlled synthetic graphs. Evaluation through a novel synthetic dataset has the potential to provide deeper insights into the comparative abilities of architectures in a stable testing environment\cite{duran2025bench}.
Despite significant advances in classification approaches, few studies combine the innate interpretability of handcrafted graph-theoretic features with diverse GNN architectures in a scalable, systematic and reproducible benchmarking framework. Our study addresses this gap by providing an evaluation of six representative GNN architectures, enriched with an optimised set of local and global features. Evaluation is performed on a novel synthetic dataset comprised of complex examples of distinct generative families, selected for their subtle topological differences.

\section{Synthetic Graph Families}\label{graph_families}
\begin{figure}[htbp]
    \centering
    \includegraphics[width=\textwidth]{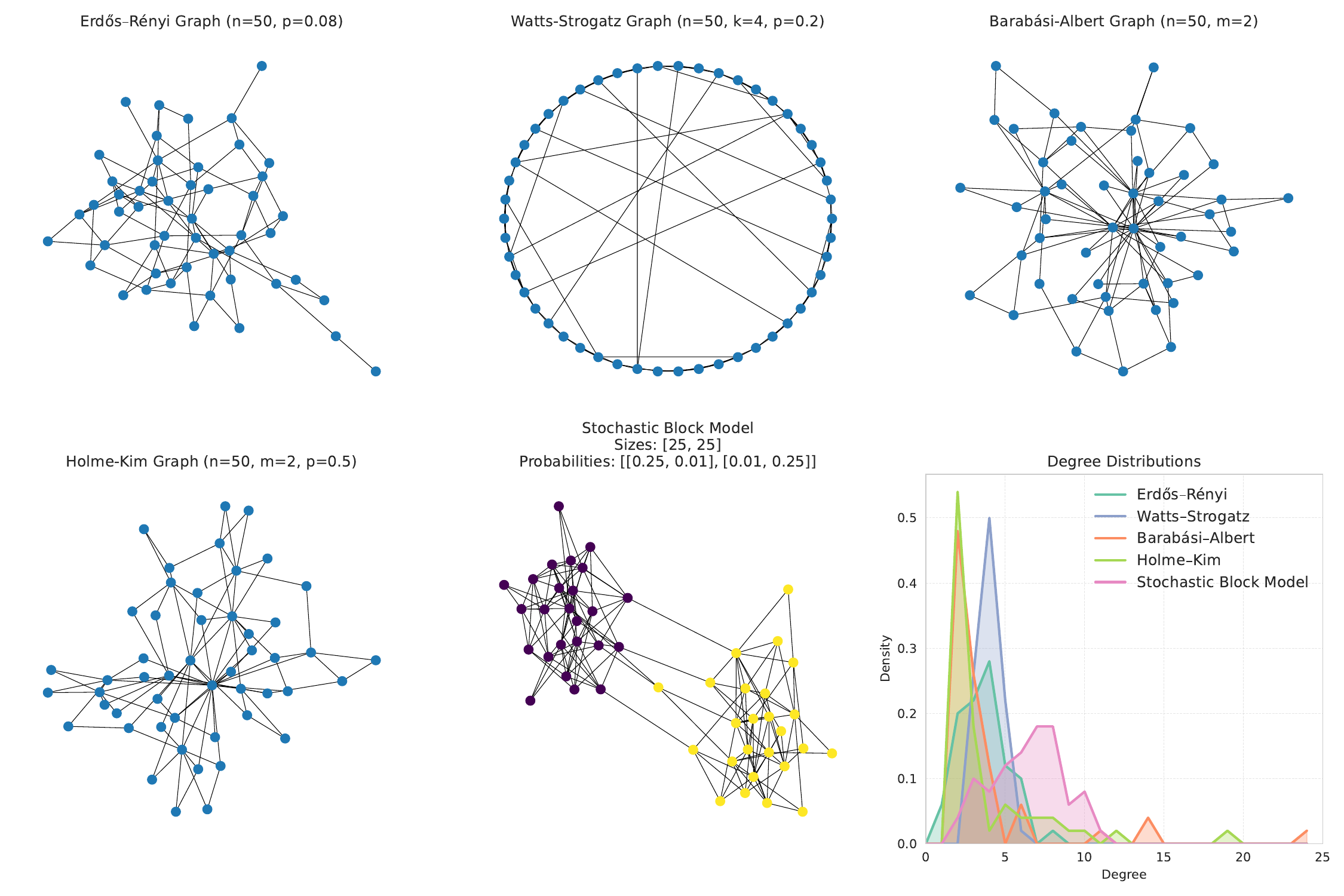}
    \caption{Visualisation of the five synthetic graph families investigated in this study, with associated degree distribution plot.}
    \label{fig:graph_types}
\end{figure}

A graph $G$ can be defined as $G(V,E)$, where $V$ and $E$ are the sets of nodes and edges in $G$ respectively. While graphs may be obtained from consideration of real-world networks, they may also be synthetically generated. Each generative function provides a distinct synthetic graph family, each with its own structural properties. Figure \ref{fig:graph_types} provides a visualisation of a representative graph from each of the five families investigated in this study, as well as a plot of their respective degree distributions. For the purposes of this figure, graph size and density have been restricted to allow for visual identification of structural variance between families.

From Figure \ref{fig:graph_types}, key structural attributes of each featured graph family may be observed. The Erdős–Rényi graph resembles an unstructured cloud, whereas the Watts-Strogatz model displays a clear ring lattice base structure. Both the Barabási-Albert and Holme-Kim graphs show clear central hubs, with the Holme Kim graph showing a tendency towards triad-based clustering. The Stochastic Block Model shows a distinct community-based structure, with strong within-block connections and fewer between block connections. Inspection of the plot of the degree distributions demonstrates distinct overlap between the Barabási-Albert and Holme-Kim examples, highlighting their structural similarities. The Watts-Strogatz and Erdős–Rényi representatives demonstrate a similar mode, albeit with differences in distribution around this value. The degree distribution of the Stochastic Block Model shows the least overlap with the other four families, demonstrating its distinct community structure. These example graphs effectively demonstrate key topological identifiers of their respective families, as well as the inherent overlap present in their structures.
\par
Each family investigated in this study is generated in such a way that they attempt to emulate some attribute of real-life networks, with some families designed specifically for improving the emulation quality of other families. The remainder of this section provides an overview of each featured graph family, including their formal mathematical constructions, and their contributions towards network modelling.
\par
One of the most well-known models for generating random graphs is the Erdős–Rényi model\cite{erdos1959graph}. While two forms of this model exist, the form that is considered in this study is the $G_{n,p}$ model, whereby for each unique node pair in $n$ nodes, an edge between them exists with equal probability $p$. This probability is independent across each unique node pair, giving an average edge number of $\binom{n}{2}p$. This can result in highly complex and dense graphs.
\par
While the simplicity of the Erdős–Rényi model lends itself to analysis, it is not necessarily directly translatable to organic networks. In particular, the two core assumptions at the heart of the Erdős–Rényi model, that all edge probabilities are equal, and that edges are independent, limit its ability to model real-world networks. For example, social networks in real life are often characterised by large, well-connected hubs. The generation of such hubs relies on interdependence and unequal probabilities that are not featured by the Erdős–Rényi model.  
Some of the limitations of the Erdős–Rényi model are addressed through the Watts-Strogatz model\cite{watts1998graph}. It takes the form $G_{n,k.p}$, where $k$ is an even integer, and $p$ is a rewiring probability. The initial graph is constructed as a ring lattice, whereby each node is connected to its $k$ nearest neighbours. Each edge is then revisited and rewired with a probability of $p$. It should be noted that although rewiring is random, the process does not allow for duplicate or multiple edges.  Through the inclusion of the rewiring process, the Watts-Strogatz model generates graphs with high clustering and short average path lengths. Watts-Strogatz graphs can be referred to as having ‘small world’ properties and provide a better approximation to certain social networks. While the Watts-Strogatz model does provide improvements on Erdős–Rényi in terms of direct translation to real-world networks, what the Watts-Strogatz model lacks is a property of many real-world systems, of being scale-free.
\par
A scale-free network has a degree distribution that follows, or asymptotically follows, a power law. This is characterized by a heavy-tailed distribution, with a small number of nodes featuring many connections. This can be seen in many real-world networks, such as a network of train stations, where a new station is more likely to be connected to well-connected stations than isolated stations, resulting in large, well-connected hubs. A generative model featuring this property is the Barabási-Albert model, $G_m$\cite{albert2002graph}. The model contains only one parameter, used to perform preferential attachment. The network is initialised with a set of disconnected nodes of size $n\geq m$. The network is then grown over time steps. At each time step, one new node is added. A set of $m$ edges is then connected to the new node, with a connection probability based on the number of existing edges attached to the neighbouring node. The probability of the new node being attached to an existing node $i$ is given as $p_i=\frac{k_i}{\sum_j k_j}$, where $k_i$ is the degree of node $i$ and the sum is taken over the degrees of all the existing nodes $j$ in the network.
\par
While the Barabási-Albert model effectively demonstrates scale-free growth and preferential attachment, it has limited clustering. This limitation was addressed through an extension to the model which allowed for the formation of triads following preferential attachment. The extension is known as the Holme-Kim model, $G_{m,p}$\cite{holme2002graph}. It is formalised by considering the preferential attachment stage of the Barabási-Albert model. After the first of the $m$ edges is connected via preferential attachment, a decision is made for each of the remaining $m-1$ edges. With probability $p$, a 'triad formation' step occurs, where an edge is added to a neighbour of the node just connected to. Otherwise, with probability $1-p$, a standard preferential attachment step is performed. In this way, Holme-Kim graphs exhibit clustering, as well as scale free growth and preferential attachment. 
\par
An alternative modelling approach involves the explicit formation of community structures through probabilistic connections between node groups. This is demonstrated by the Stochastic Block Model\cite{holland1983sbm}, which effectively represents networks characterized by modular structures, such as social communities. The Stochastic Block Model can be defined as $G_{n,C,P}$, where $n$ is the number of nodes, $C$ is a partitioning of the set of nodes into disjoint blocks, and $P$ is the symmetric probability matrix that defines probabilities for edges existing within and between blocks. It should be noted that if $P$ is constant, the resultant graph is Erdős–Rényi. 
\par
When considered in terms of the model's generation, the five graph families detailed in the section offer significant structural overlap, making their accurate classification a non-trivial task that requires consideration of global and local patterns. Each model effectively captures elements of real-world networks, with varying levels of scope for complexity and parameter sensitivity.

\section{Dataset Generation}\label{data_gen}
We generated the graphs for the dataset using the igraph library\cite{csardi2005igraph} due to its scalability and inbuilt generative functions for synthetic families, with the capacity for sufficient tuning of generative parameters such that a diverse dataset could be compiled. The choice of library was especially relevant in the context of dense Erdős–Rényi graphs, which proved unwieldy to produce at scale with alternative libraries. Due to igraph not supporting a generative function for Holme-Kim graphs, these were generated using the networkx library\cite{hagberg2008networkx} and then converted to igraph format. The igraph library also provides optimised functions for obtained useful graph-theoretic features from the generated networks, promoting computational efficiency.

To construct a dataset for evaluating classification performance on the generative models detailed in section \ref{graph_families}, a set of 400 graphs was generated for each family. The uniform number of class instances across families ensured a balanced dataset. Graphs were generated with initial node counts sampled from a uniform distribution with bounds of $5\times10^3$ and $1\times10^4$. Once generated, graphs were screened for connectivity. Any disconnected graphs were replaced with their largest connected subgraph. As a result, the final graph dataset featured a node range of $5000-9996$, with edge range of $13260-111643$. This range represents a structurally diverse dataset suitable for benchmarking at scale. A summary of the graph size ranges across each generative family is provided in Table \ref{tab:graph_sizes}:

\begin{table}[h]
\caption{Graph Family Sizes}\label{tab:graph_sizes}
\begin{tabular*}{\textwidth}{@{\extracolsep{\fill}}lp{1cm}p{1cm}p{1cm}p{1cm}p{1cm}p{1cm}@{}}
\toprule
\textbf{Family}&\textbf{Max.\newline Nodes}&\textbf{Min.\newline Nodes}&\textbf{Avg.\newline Nodes}&\textbf{Max.\newline Edges}&\textbf{Min.\newline Edges}&\textbf{Avg.\newline Edges}\\
\midrule
Erdős–Rényi           & 9989& 5010& 7656& 72242& 13260& 37988\\
Barabási-Albert       & 9995& 5000& 7463&  49960&  14994& 29439\\
Watts-Strogatz        & 9989& 5005& 7609&  59934& 20112& 37708\\
Stochastic Block Model& 9972& 5000& 7497& 111643& 15240& 46743\\
Holme-Kim             & 9996& 5002& 7518&  49931&  15486& 29546\\
\botrule
\end{tabular*}
\end{table}
\section{Feature Engineering and Selection}\label{feat_select}
To improve interpretability and reduce final model complexity, a hybrid feature-based approach was used. We engineered a set of graph-theoretic features at both the node and graph levels. These features were then systematically evaluate for their ability to provide class separation between the featured generative families, using a recursive feature pruning pipeline.
Feature selection was performed by Random Forest Classifier (RFC) models. These models were selected due to their robustness to feature scaling and their inbuilt feature importance methods\cite{breiman2001randomforests}. In each pass, to ensure the stability of the feature importance scores and mitigate variance from any single model's random initialisation, an ensemble of 5 RFC models was created and fitted to the feature values and class labels. The mean feature importance was taken from these models, and the features with the highest importance up to a cumulative importance of 0.8 were retained. This threshold was chosen to retain the most informative features while minimising redundancy and promoting interpretability. Recursive passes continued until no features were removed in a pass. The node and graph level features retained following pruning are detailed in Table \ref{tab:features_retained}:

\begin{table}[h]
\caption{Features Retained Following Pruning}\label{tab:features_retained}
\begin{tabular*}{\textwidth}{@{\extracolsep{\fill}}llp{7cm}@{}}
\toprule
\textbf{Feature}&\textbf{Level}&\textbf{Description}\\
\midrule
Eigenvector Centrality&Node&Influence of a node, scored in terms of connections to well connected nodes\\
Degree&Node&Number of connections between a node and all other nodes\\
Closeness Centrality&Node&Reciprocal of the sum of the length of the shortest paths between a node and all other nodes\\
Degree Variance&Graph&Global variance of node degrees\\
Clustering&Graph&Measurement of the connectivity of the neighbours of a node\\
Assortativity&Graph&Measure of the similarity between connected nodes\\
\botrule\\
\end{tabular*}
\end{table}

\begin{figure}[h]
     \centering
     \begin{subfigure}[b]{\textwidth}
         \centering
         \includegraphics[width=\textwidth]{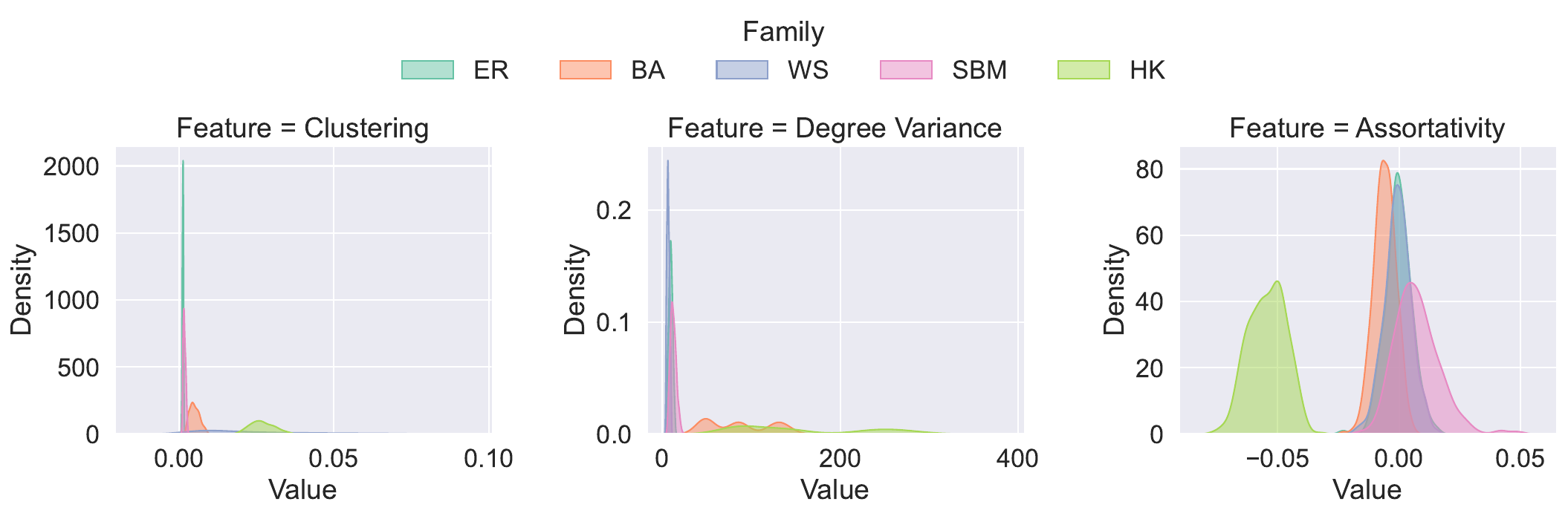}
     \end{subfigure}
     \vspace{2em}
     \begin{subfigure}[b]{\textwidth}
         \centering
         \includegraphics[width=\textwidth]{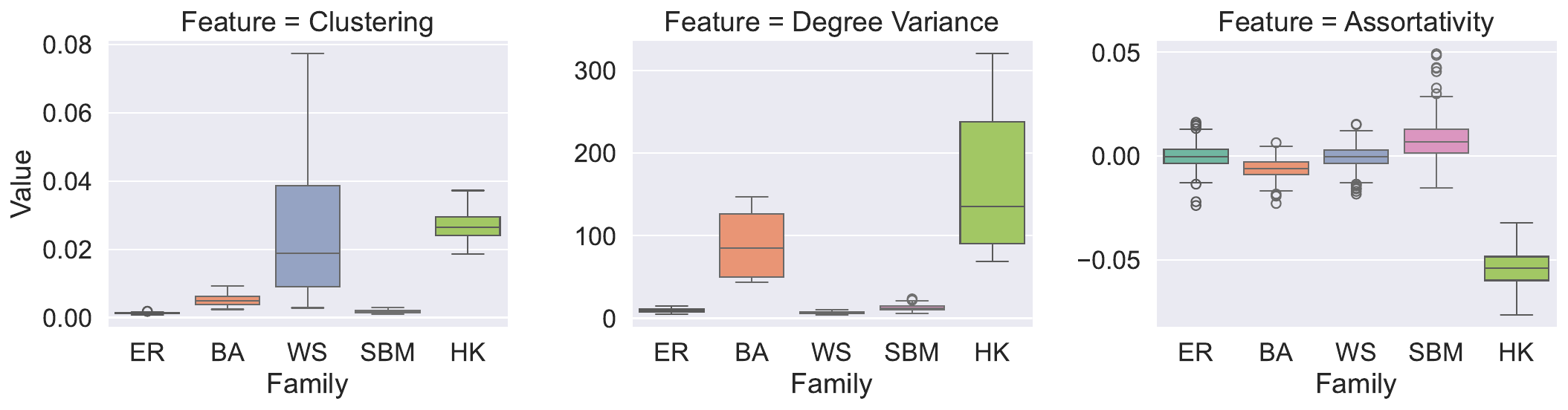}
     \end{subfigure}
     \caption{KDE and box-plots of graph-level feature distributions}
    \label{fig:graph_eda}
\end{figure}

\begin{figure}[h]
     \centering
     \begin{subfigure}[b]{\textwidth}
         \centering
         \includegraphics[width=\textwidth]{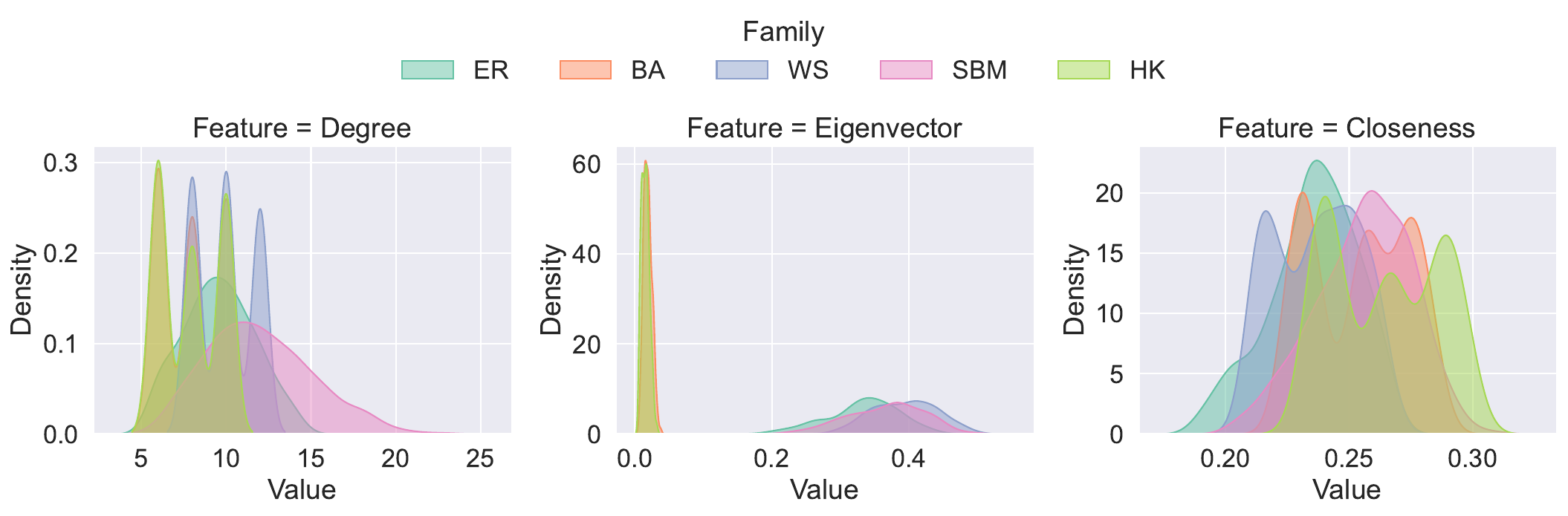}
     \end{subfigure}
     \vspace{2em}
     \begin{subfigure}[b]{\textwidth}
         \centering
         \includegraphics[width=\textwidth]{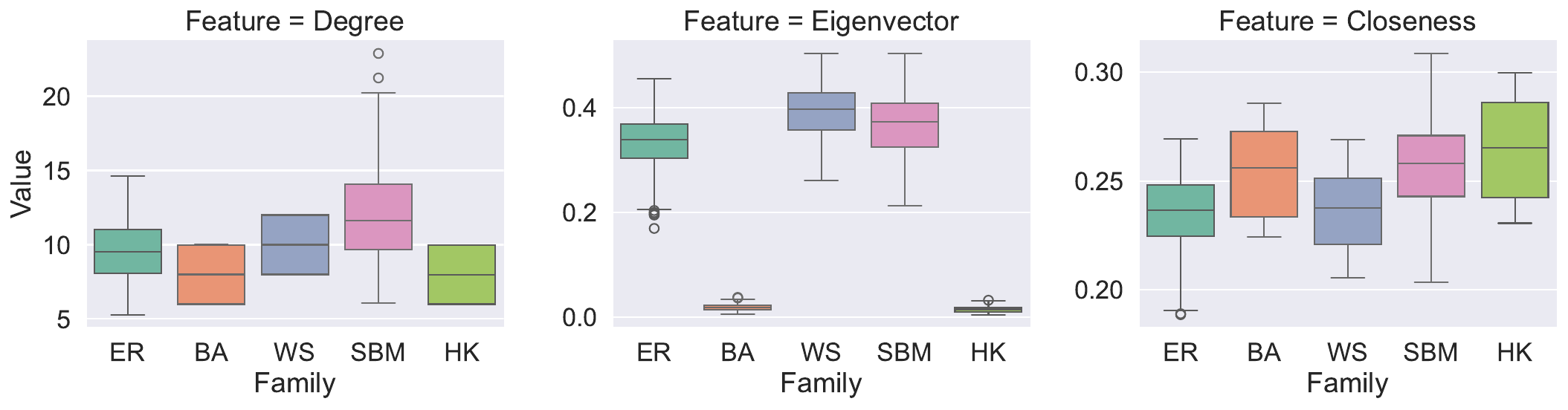}
     \end{subfigure}
     \caption{KDE and box-plots of node-level feature distributions}
    \label{fig:node_eda}
\end{figure}

To validate the pruning process, visualisations in the form of KDE and box plots were obtained. These visualisations, provided in Figures \ref{fig:graph_eda} and \ref{fig:node_eda} demonstrate the separation of each of the network families achieved by the retained features.

Inspection of Figures \ref{fig:graph_eda} and \ref{fig:node_eda} shows that one single feature does not provide good discrimination between all families. In some cases, such as when considering Assortativity, only one family is clearly separated from the others. It is evident that a set of features is required to achieve meaningful separation between each class. 
The box-plots in Figures \ref{fig:graph_eda} and \ref{fig:node_eda} illustrate class-based variations in feature distributions. The plots offer insight into the structural similarities between certain classes and confirm that the graphs generated to represent the distinct families reflect their expected traits. For example, the degree variance of the Barabási–Albert and Holme-Kim families are noticeably larger than those of the other families, reflected the preferential attachment methods involved in their generation. To further validate this, and to assess potential redundancy in feature selection, correlation heatmaps were obtained for the local and global features. In the case of local features, the mean of the feature vector for each graph was used in the correlation. The heatmaps can be seen in Figures \ref{fig:graph_feat} and \ref{fig:node_feat}:

\begin{figure}[H]
     \centering
     \begin{subfigure}[b]{0.45\textwidth}
         \centering
         \includegraphics[width=\textwidth]{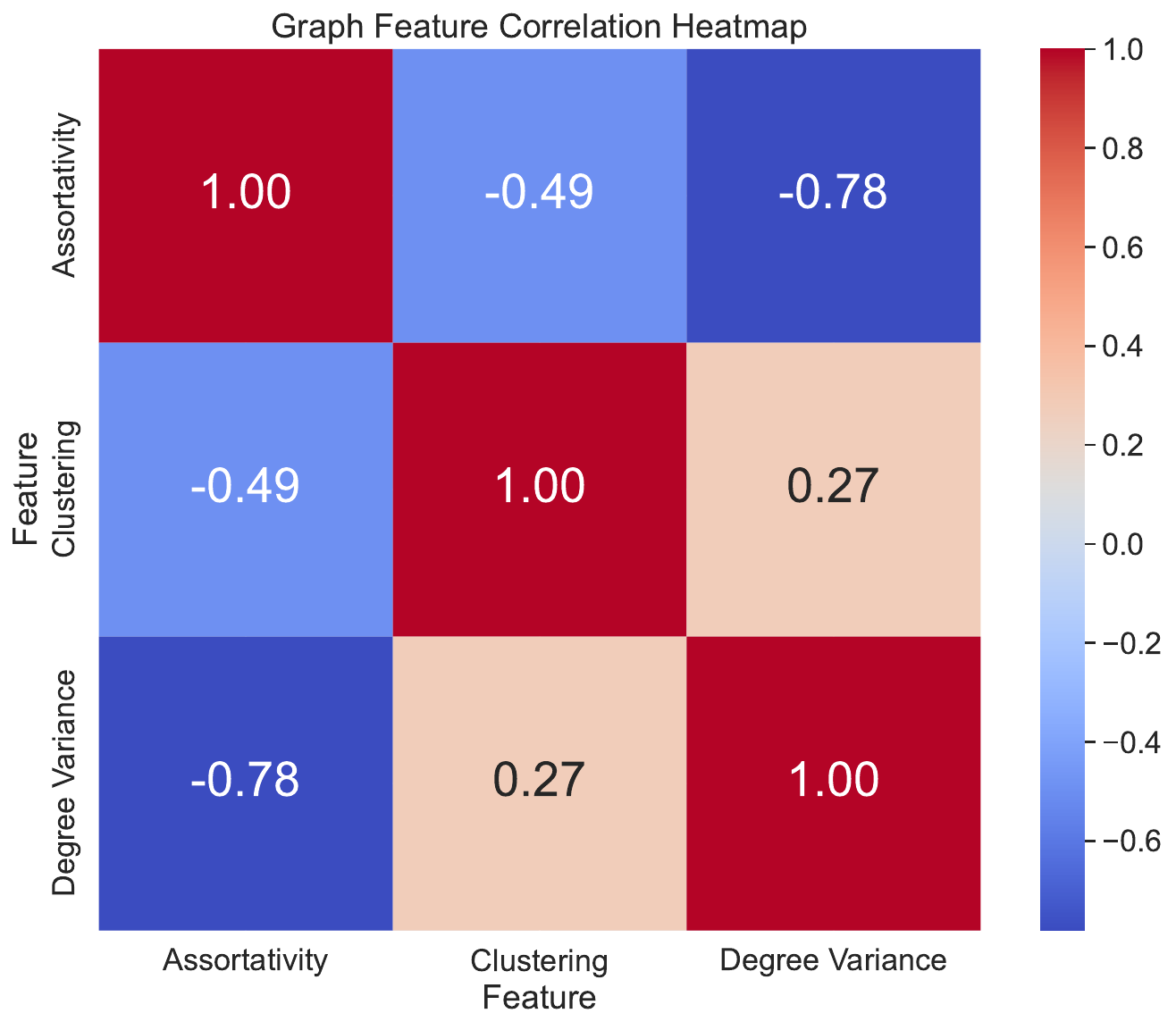}
         \caption{Graph-level features}
         \label{fig:graph_feat}
     \end{subfigure}
     \hfill
     \begin{subfigure}[b]{0.45\textwidth}
         \centering
         \includegraphics[width=\textwidth]{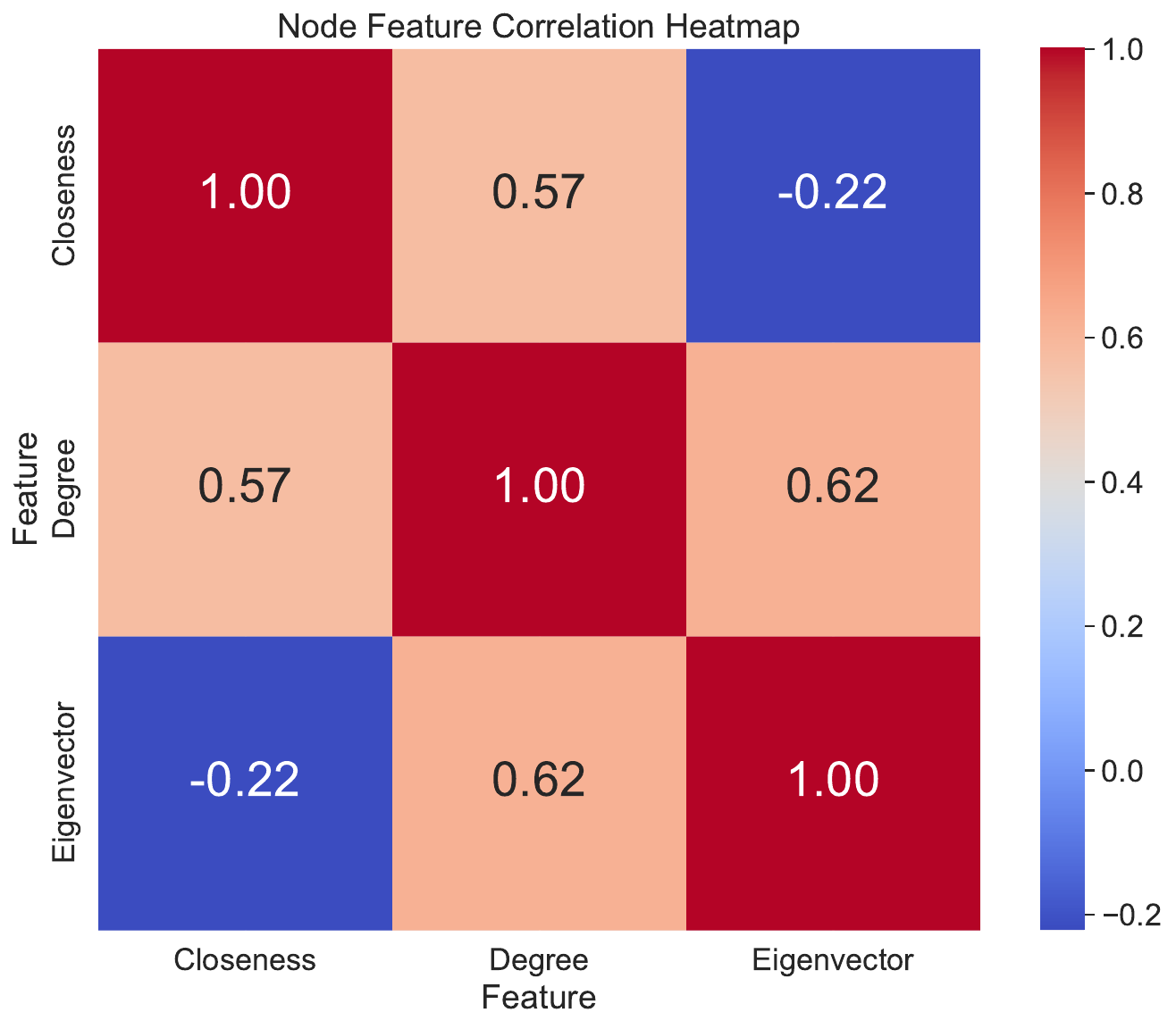}
         \caption{Node-level features}
         \label{fig:node_feat}
     \end{subfigure}
     \caption{Correlation heatmaps for (a) graph-level features and (b) node-level features.}

\end{figure}

From the heatmaps of the local and global features, moderate correlations are observed, with no features demonstrating perfect collinearity. Given that each feature represents a distinct structural attribute of graphs, the observed correlations were determined not to reflect feature redundancy, and all were carried forward to the final models. 

\section{Candidate Models}\label{cand_models}
This section outlines the baseline and GNN models evaluated in the study, including an overview of architectural structures, mathematical formulation of core operations, and the integration of engineered graph-theoretic features.

To establish a performance baseline against which to compare the GNN architectures, we implemented a Support Vector Machine (SVM) model, trained and fitted to the sets of engineered features and class labels. This model featured a Radial Basis Function (RBF) kernel, chosen for its ability to effectively model the non-linear class boundaries without requiring feature transformation. Fitting the model only to the optimised feature set provides a robust and interpretable baseline that helps to quantify the added value of the structural information and the message-passing operations utilised by the GNN models.

We selected six representative GNN variants for benchmarking due to their prevalence in recent literature, as well as their diversity in aggregate and update mechanisms. Each architecture represents a distinct aggregation or update strategy, thereby enabling systematic comparison within a unified optimisation and training framework.

Before detailing each architecture, we first formalise the common notation used. All models operate on a graph defined as as $G=(V,E)$, with $n=|V|$ nodes. We define the input node feature matrix as  $\textbf{X}\in \mathbb{R}^{n\times d}$, where $d$ is the number of features per node.

 Graph Convolutional Networks (GCN) specifically use a constant average aggregation method across neighbourhoods using the normalised adjacency matrix to update node embeddings:
 
%  \begin{equation}
% \textbf{X}' = \hat{\textbf{D}}^{-1/2}\hat{\textbf{A}}\hat{\textbf{D}}^{-1/2}\textbf{X}\mathbf{\Theta}
% \end{equation}

\begin{equation}\label{eq:gcn-propagation}
  \mathbf{X}'
    = \hat{\mathbf{D}}^{-\frac12}\,
      \hat{\mathbf{A}}\,
      \hat{\mathbf{D}}^{-\frac12}\,
      \mathbf{X}\,
      \mathbf{\Theta}
\end{equation}

where $\hat{\textbf{A}}$ is the adjacency matrix with self-loops, $\hat{\textbf{D}}$ is the degree matrix, and $\mathbf{\Theta}$ is a trainable weight matrix. This yields an updated node embeddings matrix, wherein each row $x_i$ corresponds to the updated representation of node $i$.

In contrast to GCN's static neighbourhood averaging, Graph Attention Networks (GAT), and their updated variant GATv2 use static and dynamic attention mechanisms to weight neighbourhood aggregations. The node embedding for node $i$ can be found as:

\begin{equation}\label{eq:gat-propagation}
  x'_i
    = \sigma\!\Bigl(
        \sum_{j \in \mathcal{N}(i)}
          \alpha_{ij}\,\mathbf{\Theta}\,x_j
      \Bigr)
\end{equation}

where $\mathbf{\Theta}$ is a shared weight matrix, and $a_ij$ is the attention vector between nodes $i$ and $j$. The key difference between GAT and GATv2 lies in the formulation of the attention vectors. In standard GAT models, the vector $\alpha_{ij}$ is found as: 

% \[
% \alpha_{ij} = \text{softmax}_j\left( \text{LeakyReLU} \left( \mathbf{a}^\top [\mathbf{\Theta} x_i \, \| \, \mathbf{\Theta} x_j] \right) \right)
% \]

\begin{equation}\label{eq:gat-attention}
  \alpha_{ij}
    = \operatorname{softmax}_j\Bigl(
        \operatorname{LeakyReLU}\bigl(
          \mathbf{a}^\top\,[\,\mathbf{\Theta}\,x_i \;\Vert\; \mathbf{\Theta}\,x_j\,]
        \bigr)
      \Bigr)
\end{equation}

where $||$ denotes concatenation. In GATv2, this is altered so that the weight matrix is applied following concatenation, giving:

% \[
% \alpha_{ij}^{\text{v2}} = \text{softmax}_j\left( \text{LeakyReLU} \left( \mathbf{a}^\top \mathbf{\Theta} [x_i \, \| \, x_j] \right) \right)
% \]

\begin{equation}
\alpha_{ij}^{\text{v2}} = \text{softmax}_j\Bigl(
    \text{LeakyReLU} \left(
        \mathbf{a}^\top \mathbf{\Theta} [x_i \, \| \, x_j]
    \right) \Bigr)
\label{eq:attention-coeff}
\end{equation}

This allows a more expressive and context dependent weighting of neighbours, allowing for attention weights to depend explicitly on the query node $i$ itself.

Meanwhile, Graph Sample and Aggregate (GraphSAGE) takes an inductive approach to sample neighbourhoods, aggregating using mean:

% \[x'_i = \sigma\left(\mathbf{\Theta}\cdot\left[x_i||\frac{1}{|N(i)|}\sum_{j\in N(i)}x_j\right]\right)\]

\begin{equation}
x_i' = \sigma\left( \boldsymbol{\Theta} \cdot \left[ x_i \, \| \, \frac{1}{|\mathcal{N}(i)|} \sum_{j \in \mathcal{N}(i)} x_j \right] \right)
\label{eq:node-update}
\end{equation}

The inductive modelling approach allows for stable generalisation to unseen localities, as well as efficient computation, as only a sample of each neighbourhood is processed at node updates.

In contrast to the mean-based aggregation approaches of GCN, GAT, and SAGE, Graph Isomorphism Networks (GIN) use a sum aggregation that is passed to a Multilayer Perceptron for updating, using the additional retained information to achieve equivalent discriminative power to the Weisfeiler-Lehman test:

% \[x'_i = \text{MLP}\left((1+\epsilon)\cdot x_i+\sum_{j\in N(i)}x_j\right)\]

\begin{equation}
x_i' = \text{MLP} \left( (1 + \epsilon) \cdot x_i + \sum_{j \in \mathcal{N}(i)} x_j \right)
\label{eq:gin-update}
\end{equation}

where $\epsilon$ is either a learnable parameter or a fixed scalar. In this study, the variable $\texttt{train\_eps}$ is set to True, allowing for the value of $\epsilon$ to be learned. The use of sum-based aggregation vs mean or attention helps to improve the models ability to preserve overall structure.

Unlike the other models discussed in this section, which update node embeddings based on localities, Graph Transformer Networks (GTN) use a global attention field to capture long range hierarchies, allowing for deep insight into global structures:

% \[x'_i = \sum_{j=1}^na_{ij}\mathbf{\Theta}x_j\]

\begin{equation}
x_i' = \sum_{j=1}^{n} a_{ij} \, \boldsymbol{\Theta} \, x_j
\label{eq:linear-propagation}
\end{equation}

The global attention functionality is illustrated in comparison of the sum limits between GTN and GAT models. The attention vector $\alpha_{ij}$ is found as:

% \[
% \alpha_{ij} = \text{softmax}_j\left( \frac{(\mathbf{\Theta}_Q x_i)^\top (\mathbf{\Theta}_K x_j)}{\sqrt{d}} \right)
% \]

\begin{equation}
\alpha_{ij} = \mathrm{softmax}_j\left( \frac{ \left( \boldsymbol{\Theta}_Q x_i \right)^\top \left( \boldsymbol{\Theta}_K x_j \right) }{ \sqrt{d} } \right)
\label{eq:attention-score}
\end{equation}

where $\mathbf{\Theta}_Qx_i$ and $\mathbf{\Theta}_Kx_j$ are the query vector for node $i$ and the key vector for node $j$ respectively.

Having defined the core mathematical operations for each GNN variant featured in this study, we now describe their shared implementation details. All models were created using PyTorch Geometric and were trained and optimised under matched pipeline conditions.

Each model was provided with the adjacency matrix of each graph, as well as the sets node and graph level features. To ensure compatibility across GNN variants, all models in this study share a base architectural scaffold. Each model consists of two architecture-specific convolutional layers, with each convolutional layer followed by batch normalisation, dropout and global mean pooling.

The forward pass of each model can be described as follows:
Node features and adjacency matrix information are passed through two GNN convolutional layers, enabling aggregation of information across node localities. Each convolution is followed by ReLU activation, batch normalisation and dropout. Bath Normalisation is applied following both convolutional and fully connected stages to provide stability across models and graph sizes. Dropout layers mitigate against overfitting, a common challenge in deep models such as GATv2 and GTN.

The output of the final convolution	layer is aggregated via global mean pooling to produce an embedding for each sample in the training batch prior to combination with the graph-level features. Global mean pooling offers a simple approach to aggregation that is compatible across architectures. The concatenated embedding is passed through a fully connected layer with ReLU activation, followed by batch normalisation and dropout, before a final linear output layer, producing class logits. Alongside the class logits, the model returns a copy of the embeddings stored prior to the output layer, supporting downstream visualisation for interpretability and analysis.

By developing a standardised architectural scaffold and isolating the aggregation mechanisms, we ensure that differences in performance can be attributed to the strengths of the given mechanisms, rather than confounding factors in training procedure or model complexity.

\section{Training and Optimisation}\label{training_optim}
All models were compiled using an Adam optimiser and a learning rate scheduler that reduced the learning rate by a factor of 0.5, with patience of 5. This allowed for a smoother convergence once performance metrics reached plateaus in training. All models used cross-entropy loss for the criterion. Given the context of a classification task for discrete classes using a balanced dataset, it was deemed unnecessary to explore the use of weighted cross-entropy or focal loss.

To find the optimal hyperparameters for each architecture for this training regimen, we used Optuna\cite{akiba2019optuna} to perform a systematic search. For each architecture, 50 trials we run to establish an optimal combination of hidden channels, learning rate, and dropout rate for each architecture, using validation loss as the comparison metric. A combined search space was used to inform hyperparameter ranges for each GNN architecture. While individual architectures may benefit from more tailored search spaces, the ranges featured for each variable allowed for flexibility and scalability in architectural choices.

The range of hidden channels was constrained to 32, 64, 96, and 128. The learning rate was constrained to a float between $1e-5$ and $3e-3$, and the dropout rate was constrained to a float between 0.2 and 0.5. These ranges of hyperparameters, combined with the three model architectures provided a robust selection of model tunings for testing.  
The tuning options are summarised below, in Table \ref{tab:tuning_params}:

\begin{table}[h]
\caption{Tuning options for Optuna trialling}\label{tab:tuning_params}
\begin{tabular*}{\textwidth}{@{\extracolsep{\fill}}ll@{}}
\toprule
\textbf{Variable}&\textbf{Range}\\
\midrule
Architecture&[GCN, GIN, GAT, GATv2, SAGE, GTN]\\
Hidden Channels&[32,64,96,128]\\
Initial Learning Rate&($1e-5,\;3e-3)$\\
Dropout Rate&($0.2,\;0.5)$\\
\botrule\\
\end{tabular*}
\end{table}

During hyperparameter tuning, trial models were trained over 10 epochs to identify promising trial combinations. Also tracked during the tuning process were the time taken to train across the 10 epochs and the number of trainable parameters for each model.
Validation loss was used as the metric for ranking candidate models, with validation accuracy additionally reported for further insight into the discriminative power of the models.

Following completion of the trialling process, the best performing candidate models of each architecture type were tabulated for comparison, as below. Table \ref{tab:candidate_models} presents the optimised variables for each candidate architecture, and Table \ref{tab:candidate_model_metrics} illustrates the training time, model size in terms of trainable parameters, and performance metrics captured for each model.
\begin{table}[htbp]
\caption{Optimised tuning parameters for each architecture.}\label{tab:candidate_models}
\begin{tabular*}{\textwidth}{@{\extracolsep{\fill}}llll@{}}
\toprule
\textbf{Architecture}&\textbf{Hidden Channels}&\textbf{Learning Rate}&\textbf{Dropout Rate}\\
\midrule
GTN & 96  & 0.001761907 & 0.415116501 \\
SAGE & 128 & 0.002951596 & 0.39580436 \\
GATV2 & 128 & 0.001253465 & 0.389288295 \\
GIN & 96  & 0.002046029 & 0.439854549 \\
GCN & 96  & 0.001427652 & 0.202616084 \\
GAT & 128 & 0.002977527 & 0.408811481 \\
\botrule
\end{tabular*}
\end{table}
\begin{table}[htbp]
\caption{Candidate model size, training time \& validation performance.}\label{tab:candidate_model_metrics}
\begin{tabular*}{\textwidth}{@{\extracolsep{\fill}}lllll@{}}
\toprule
\textbf{Architecture}&\textbf{Parameters}&\textbf{Training Time(s)}&\textbf{Val. Loss}&\textbf{Val. Accuracy}\\
\midrule
GTN & 49445 & 57.57614899 & 0.104280656 & 0.975 \\
SAGE & 52101 & 28.1321187 & 0.105685673 & 0.9825 \\
GATV2 & 52869 & 91.50493908 & 0.208493404 & 0.9325 \\
GIN & 38983 & 26.48632288 & 0.223608557 & 0.92 \\
GCN & 20357 & 34.32733512 & 0.249802489 & 0.9075 \\
GAT & 35845 & 54.53129792 & 0.284387849 & 0.8625 \\
\botrule
\end{tabular*}
\end{table}

The results of the Optuna trialling process provide valuable insight into the relative strengths of the architectures, with the additional context of model size and training time. Although the GTN model achieved the best performance in terms of validation loss, it is outperformed by GraphSAGE in terms of validation accuracy. Of note is that the GraphSAGE model carried out its ten epochs of training in less than half the time of that required by the GTN model. The largest model, GATv2, took significantly more time than all other models to train, and also underperformed in terms of loss and accuracy. GIN was the fastest model to train, although performance was somewhat reduced. The worse performing architectures were GCN and GAT, with GAT being the only model to achieve a validation accuracy score of $<90\%$.

Following hyperparameter tuning, a final model for each architecture was built, trained and evaluated using the optimised set of hyperparameters found during trialling. Each model was then trained for a maximum of 100 epochs, with an early stopping mechanism based on validation loss, with a patience of 10 epochs. In this way, the different architectures would be allowed to demonstrate their own rates of convergence and stability, without being constrained to a fixed number of epochs.

\section{Evaluation}\label{eval}
This section presents a comprehensive evaluation of each fully trained model. Models were assessed using accuracy, precision, recall, and F1-score. Visual summaries in the form of confusion matrices, along with t-SNE and UMAP embedding plots, are also included to support interpretability. Where appropriate, class-level performance is used to identify specific strengths and weaknesses of each architecture.

Table \ref{tab:accuracy} presents the validation accuracy scores of each fully trained model.

\begin{table}[htbp]
\centering
\captionsetup{width=0.95\textwidth}
\caption{Overall validation accuracy scores for each candidate model: visual summary (left) and numerical values (right). Models are sorted by descending accuracy.}
\label{tab:accuracy}
\begin{tabular}{@{} c c @{}}
    % Figure (left cell)
    \includegraphics[width=0.40\textwidth]{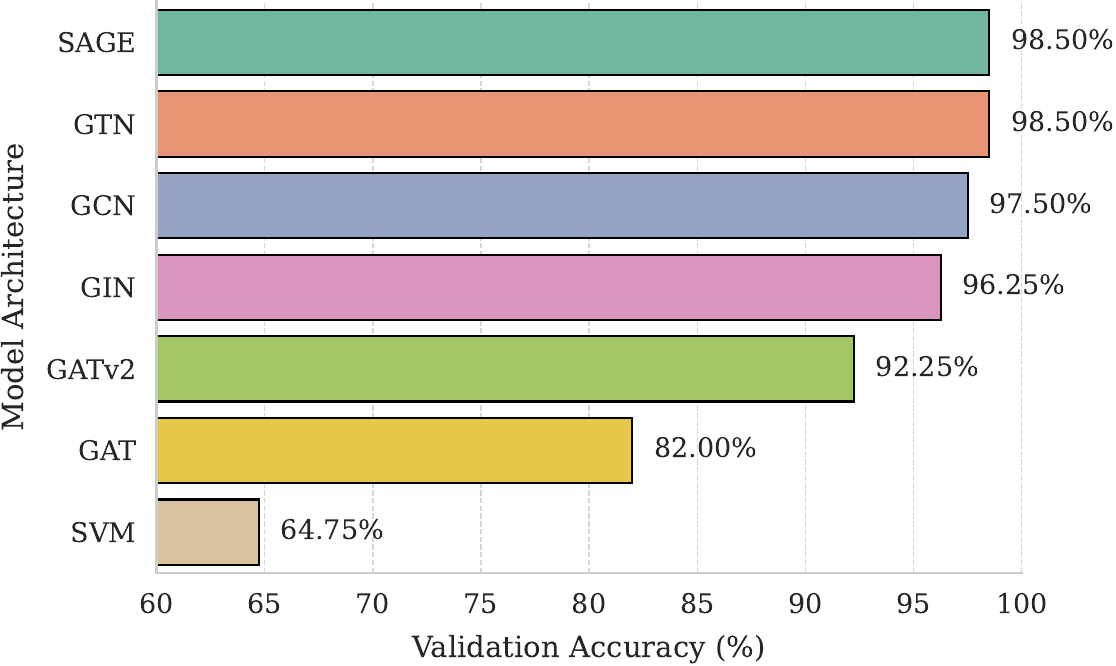}
    &
    % Table (right cell)
    \begin{tabular}[b]{l c}
        \toprule
        \textbf{Model Architecture} & \textbf{Validation Accuracy (\%)} \\
        \midrule
        SAGE     & 98.50 \\
        GTN      & 98.50 \\
        GCN      & 97.50 \\
        GIN      & 96.25 \\
        GATv2    & 92.25 \\
        GAT      & 82.00 \\
        SVM      & 64.75 \\
        \bottomrule
    \end{tabular}
\end{tabular}
\end{table}

 All GNN-based models substantially outperformed the baseline SVM model, with top performers achieving near-perfect scores. The GraphSAGE and GTN models achieved the highest overall classification accuracy, each scoring $98.50\%$. These results were followed closely by GCN ($97.50\%$) and GIN ($96.25\%$). Both of these models improved their rankings from the Optuna trials. The GAT-based models lagged behind, achieving $92.25\%$ and $82\%$ respectively. All GNN models showed significant improvements compared to the $64.75\%$ achieved by the baseline SVM model. This highlights the importance of message passing and the capacity of deep architectures to capture structural complexity.
Tables \ref{tab:precision}, \ref{tab:recall} and \ref{tab:f1score} report per-class precision, recall, and F1-Score for each architecture. 
\begin{table}[h]
\centering
\caption{Per-class precision scores across all candidate architectures.}
\label{tab:precision}
\begin{tabular*}{\textwidth}{@{\extracolsep{\fill}}lccccc|c}
\toprule
\textbf{Architecture} & \textbf{ER} & \textbf{BA} & \textbf{WS} & \textbf{SBM} & \textbf{HK} & \textbf{Mean} \\
\midrule
SAGE   & 0.9333 & 1.0000 & 1.0000 & 1.0000 & 1.0000 & 0.9867 \\
GTN    & 0.9432 & 1.0000 & 1.0000 & 0.9877 & 1.0000 & 0.9862 \\
GCN    & 0.9130 & 0.9750 & 1.0000 & 1.0000 & 1.0000 & 0.9776 \\
GIN    & 0.8485 & 1.0000 & 1.0000 & 1.0000 & 1.0000 & 0.9697 \\
GATv2  & 0.7895 & 0.9750 & 1.0000 & 0.8784 & 1.0000 & 0.9286 \\
GAT    & 0.7037 & 0.8864 & 1.0000 & 0.6034 & 1.0000 & 0.8387 \\
SVM    & 0.4493 & 0.6452 & 0.6486 & 0.6667 & 1.0000 & 0.6820 \\
\bottomrule
\end{tabular*}
\end{table}

\begin{table}[h]
\centering
\caption{Per-class recall scores across all candidate architectures.}
\label{tab:recall}
\begin{tabular*}{\textwidth}{@{\extracolsep{\fill}}lccccc|c}
\toprule
\textbf{Architecture} & \textbf{ER} & \textbf{BA} & \textbf{WS} & \textbf{SBM} & \textbf{HK} & \textbf{Mean} \\
\midrule
SAGE   & 1.0000 & 1.0000 & 1.0000 & 0.9294 & 1.0000 & 0.9859 \\
GTN    & 0.9881 & 1.0000 & 1.0000 & 0.9412 & 1.0000 & 0.9859 \\
GCN    & 1.0000 & 1.0000 & 1.0000 & 0.9059 & 0.9740 & 0.9760 \\
GIN    & 1.0000 & 1.0000 & 1.0000 & 0.8235 & 1.0000 & 0.9647 \\
GATv2  & 0.8929 & 1.0000 & 1.0000 & 0.7647 & 0.9740 & 0.9263 \\
GAT    & 0.4524 & 1.0000 & 0.9868 & 0.8235 & 0.8701 & 0.8266 \\
SVM    & 0.3875 & 1.0000 & 0.9000 & 0.5000 & 0.4500 & 0.6475 \\
\bottomrule
\end{tabular*}
\end{table}

\begin{table}[h]
\centering
\caption{Per-class F1-scores across all candidate architectures.}
\label{tab:f1score}
\begin{tabular*}{\textwidth}{@{\extracolsep{\fill}}lccccc|c}
\toprule
\textbf{Architecture} & \textbf{ER} & \textbf{BA} & \textbf{WS} & \textbf{SBM} & \textbf{HK} & \textbf{Mean} \\
\midrule
GTN    & 0.9651 & 1.0000 & 1.0000 & 0.9639 & 1.0000 & 0.9858 \\
SAGE   & 0.9655 & 1.0000 & 1.0000 & 0.9634 & 1.0000 & 0.9858 \\
GCN    & 0.9545 & 0.9873 & 1.0000 & 0.9506 & 0.9868 & 0.9759 \\
GIN    & 0.9180 & 1.0000 & 1.0000 & 0.9032 & 1.0000 & 0.9643 \\
GATv2  & 0.8380 & 0.9873 & 1.0000 & 0.8176 & 0.9868 & 0.9260 \\
GAT    & 0.5507 & 0.9398 & 0.9934 & 0.6965 & 0.9306 & 0.8222 \\
SVM    & 0.4161 & 0.7843 & 0.7539 & 0.5714 & 0.6207 & 0.6293 \\
\bottomrule
\end{tabular*}
\end{table}

Together, these results confirm that the top-performing models are capable of accurately learning from both local and global graph structures. However, subtle limitations in class-specific recall and confusion between certain graph families highlight key architectural differences. 
GTN and GraphSAGE consistently provided near-perfect results in each metric, suggesting strong generalisation and robust discrimination.GIN achieved excellent performance across most classes but exhibited a notable drop in recall for the Stochastic Block Model, where performance fell to $0.8235$. This indicates a reduced ability to distinguish modular structure when using sum-based aggregation. Each model found this family the hardest to accurately classify, with most errors occurring as a result of misclassifying Stochastic Block Models. This was emphasised when considering GATv2. Despite a strong overall F1-score of $0.9260$, the model presented a recall score of $0.7647$ for the Stochastic Block Model class. Meanwhile, the original GAT demonstrated difficulty in classifying both Erdős–Rényi and Stochastic Block Model graphs, with a mean F1-score of $0.8222$. These results reinforce the limitations of static, local attention mechanisms in GAT for capturing global graph structure. In contrast, the global attention field employed by GTN enabled stronger long-range discrimination, particularly for complex or densely connected families. The SVM model yielded a final F1-Score of $0.6293$, further demonstrating the difficulties associated with sole reliance upon handcrafted features for accurate classification of such complex networks.

Table \ref{tab:training_time} presents the number of training epochs and total training time used to train the final models.
\begin{table}[htbp]
\centering
\caption{Training duration and epochs to convergence for each architecture.}
\label{tab:training_time}
\begin{tabular*}{\textwidth}{@{\extracolsep{\fill}}lcc}
\toprule
\textbf{Architecture} & \textbf{Training Epochs} & \textbf{Training Time (s)} \\
\midrule
GTN     & 73  & 474.40 \\
GCN     & 56  & 226.50 \\
SAGE    & 49  & 167.06 \\
GATv2   & 31  & 374.73 \\
GIN     & 21  & 82.28  \\
GAT     & 7   & 92.25  \\
\bottomrule
\end{tabular*}
\end{table}

\begin{figure}[H]
\centering
\begin{subfigure}[b]{0.39\textwidth}
    \includegraphics[width=\linewidth]{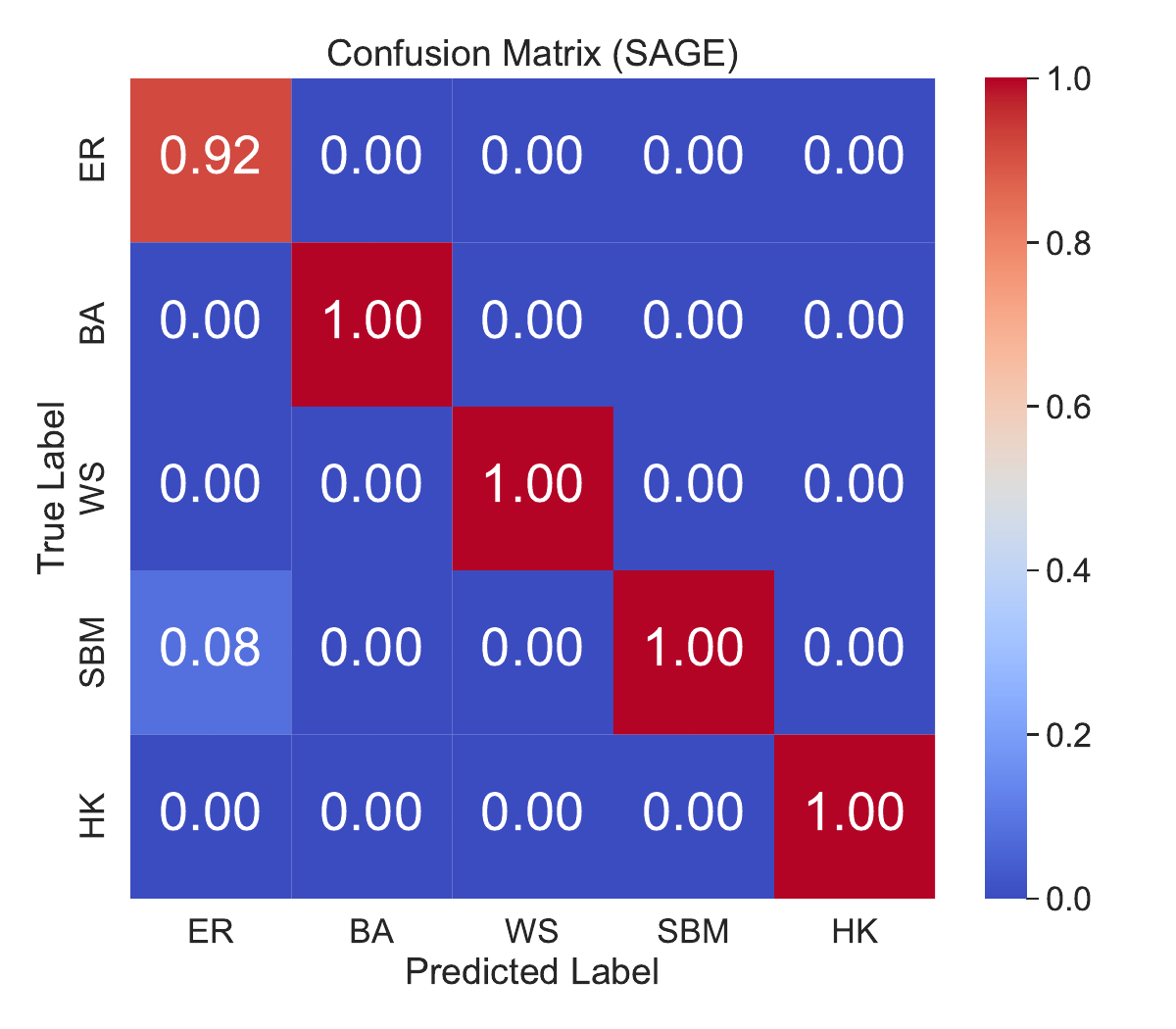}
    \caption{SAGE}
\end{subfigure}
\hfill
\begin{subfigure}[b]{0.39\textwidth}
    \includegraphics[width=\linewidth]{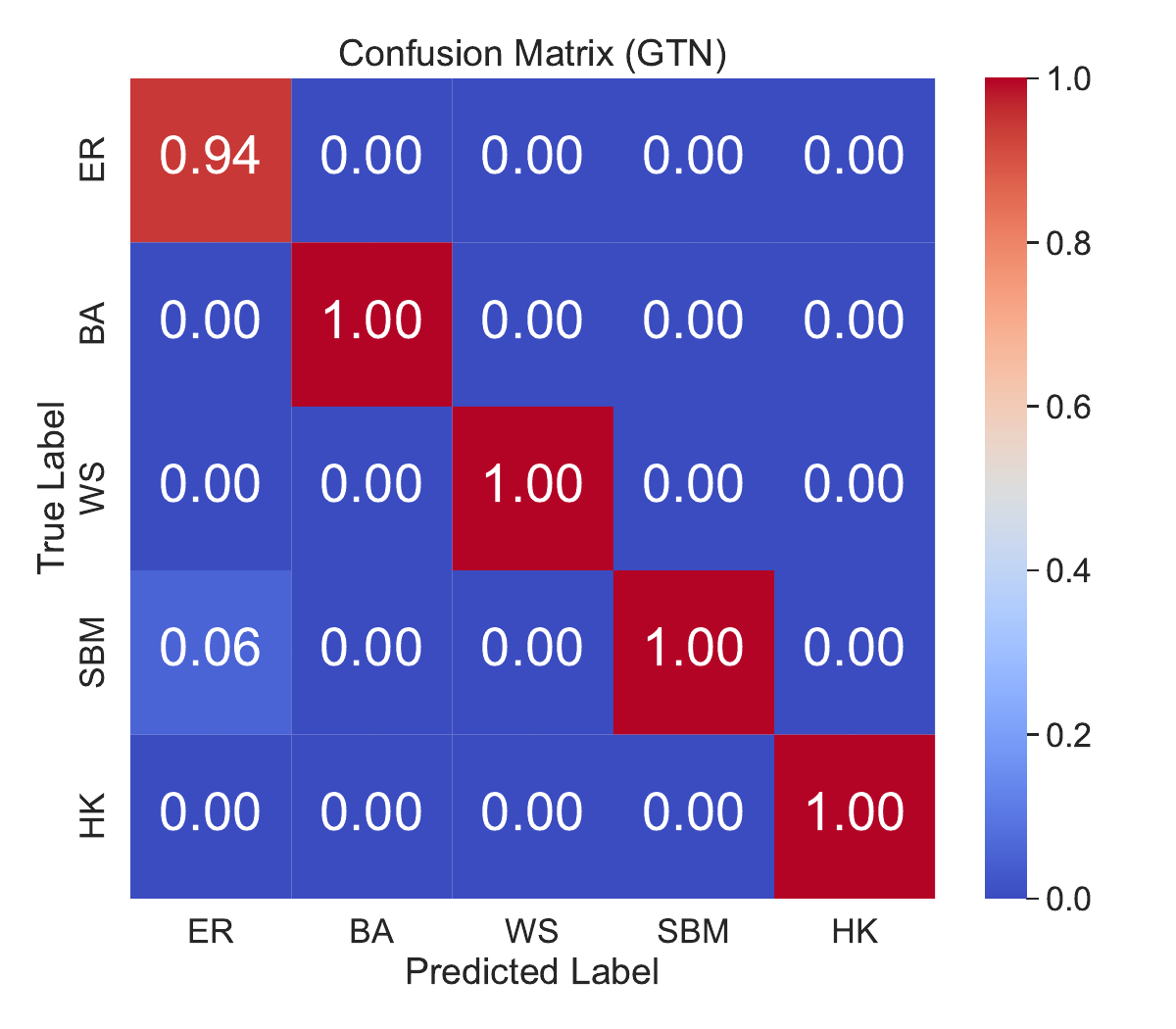}
    \caption{GTN}
\end{subfigure}

\begin{subfigure}[b]{0.39\textwidth}
    \includegraphics[width=\linewidth]{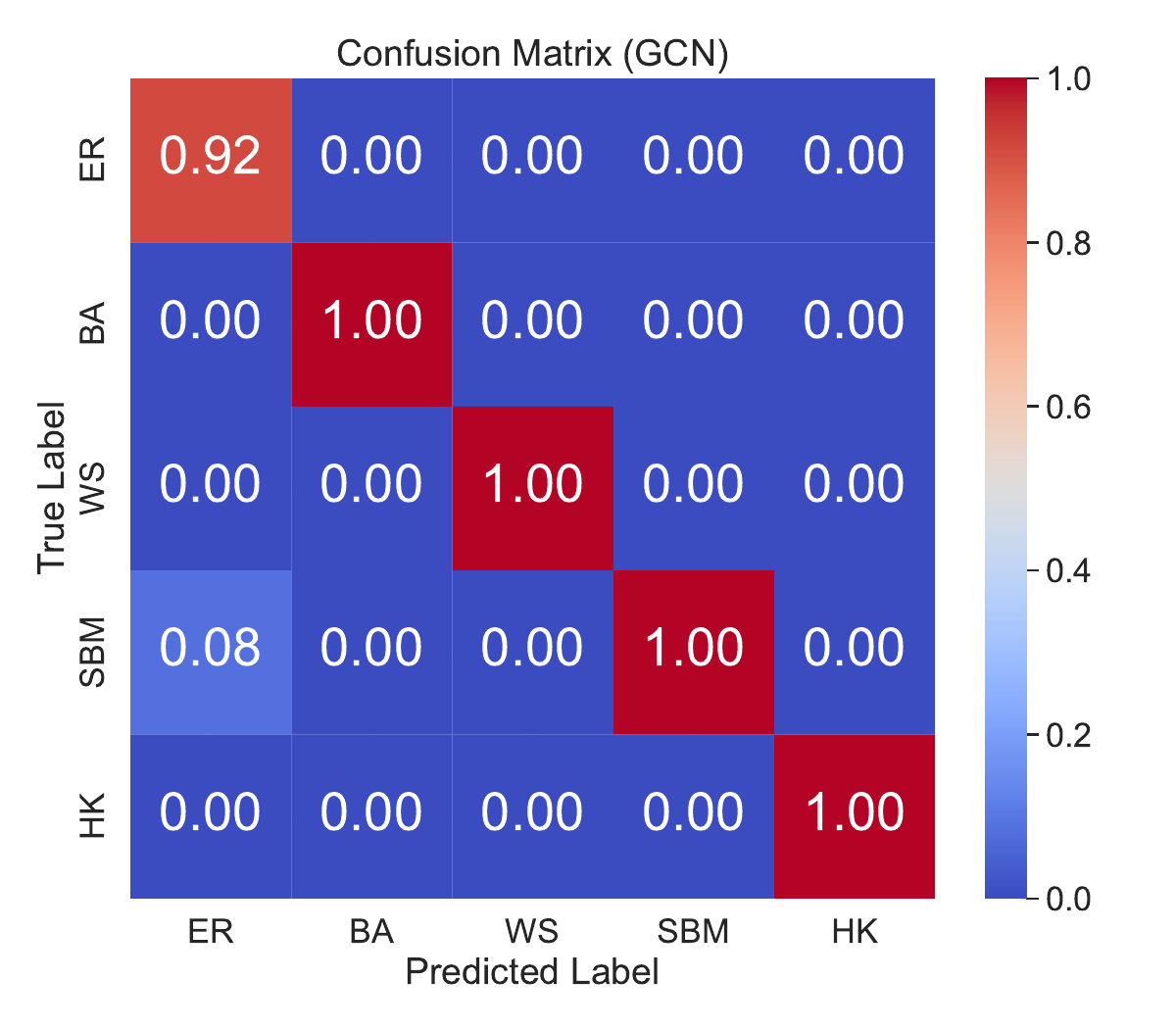}
    \caption{GCN}
\end{subfigure}
\hfill
\begin{subfigure}[b]{0.39\textwidth}
    \includegraphics[width=\linewidth]{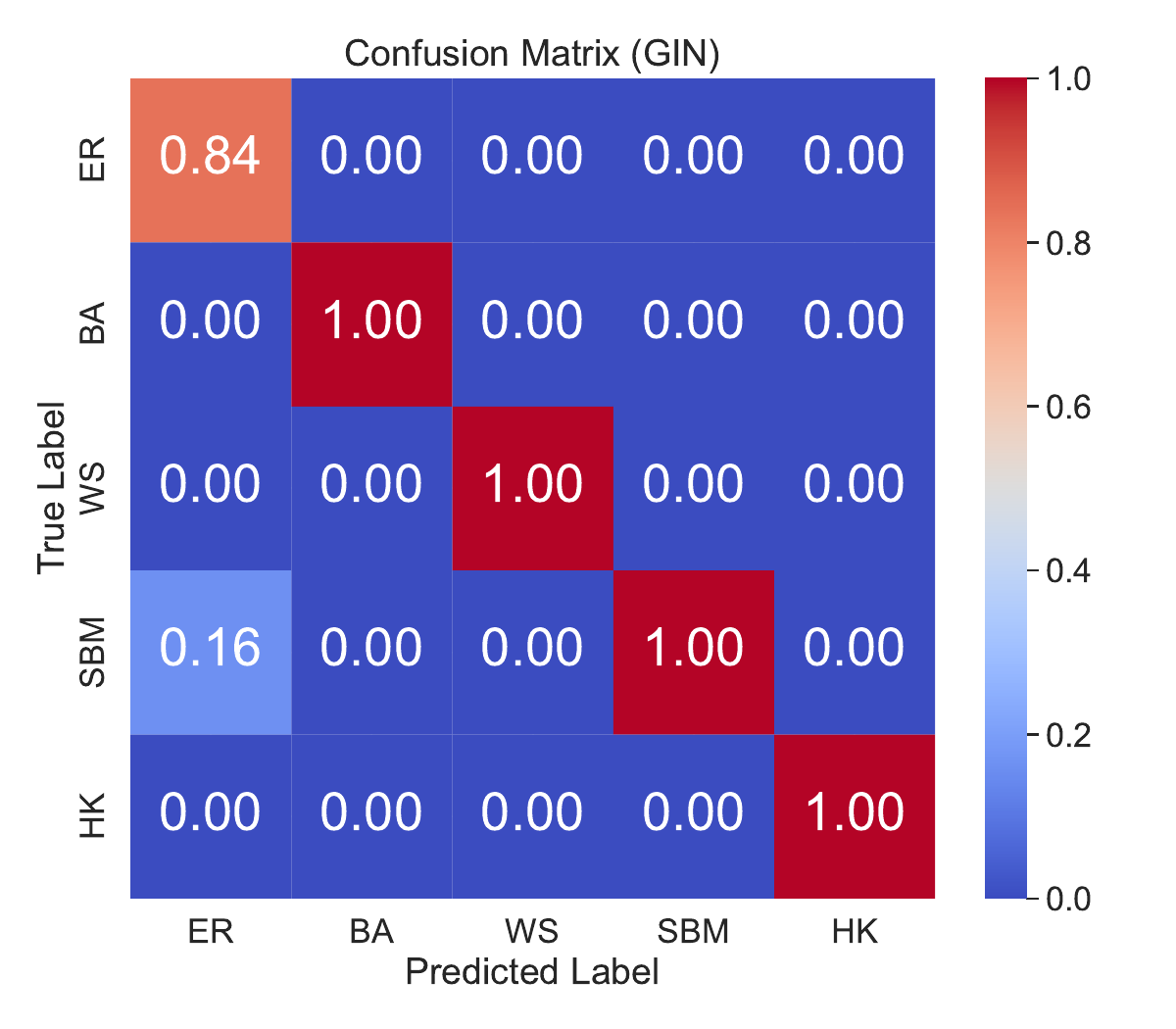}
    \caption{GIN}
\end{subfigure}

\begin{subfigure}[b]{0.39\textwidth}
    \includegraphics[width=\linewidth]{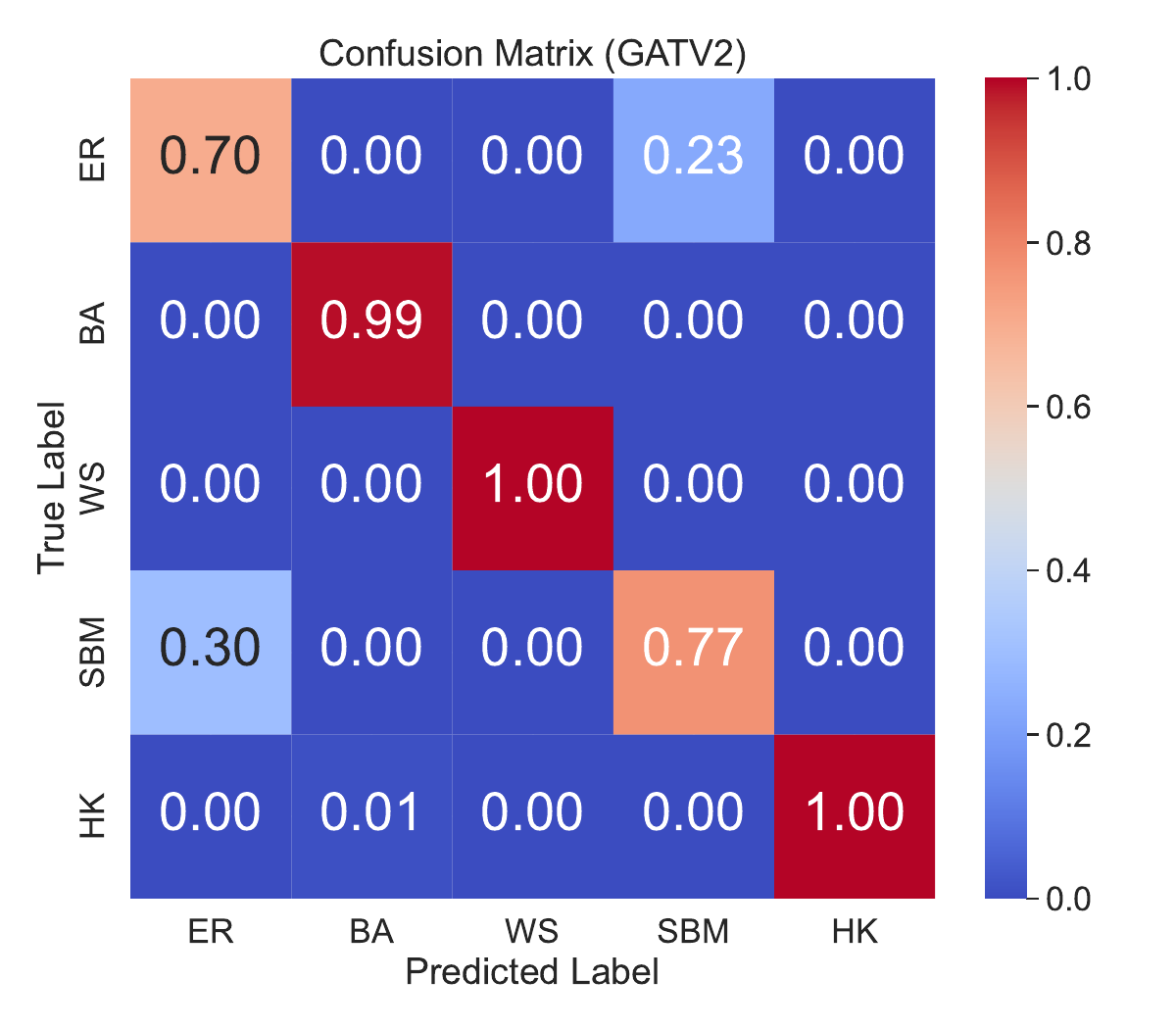}
    \caption{GATv2}
\end{subfigure}
\hfill
\begin{subfigure}[b]{0.39\textwidth}
    \includegraphics[width=\linewidth]{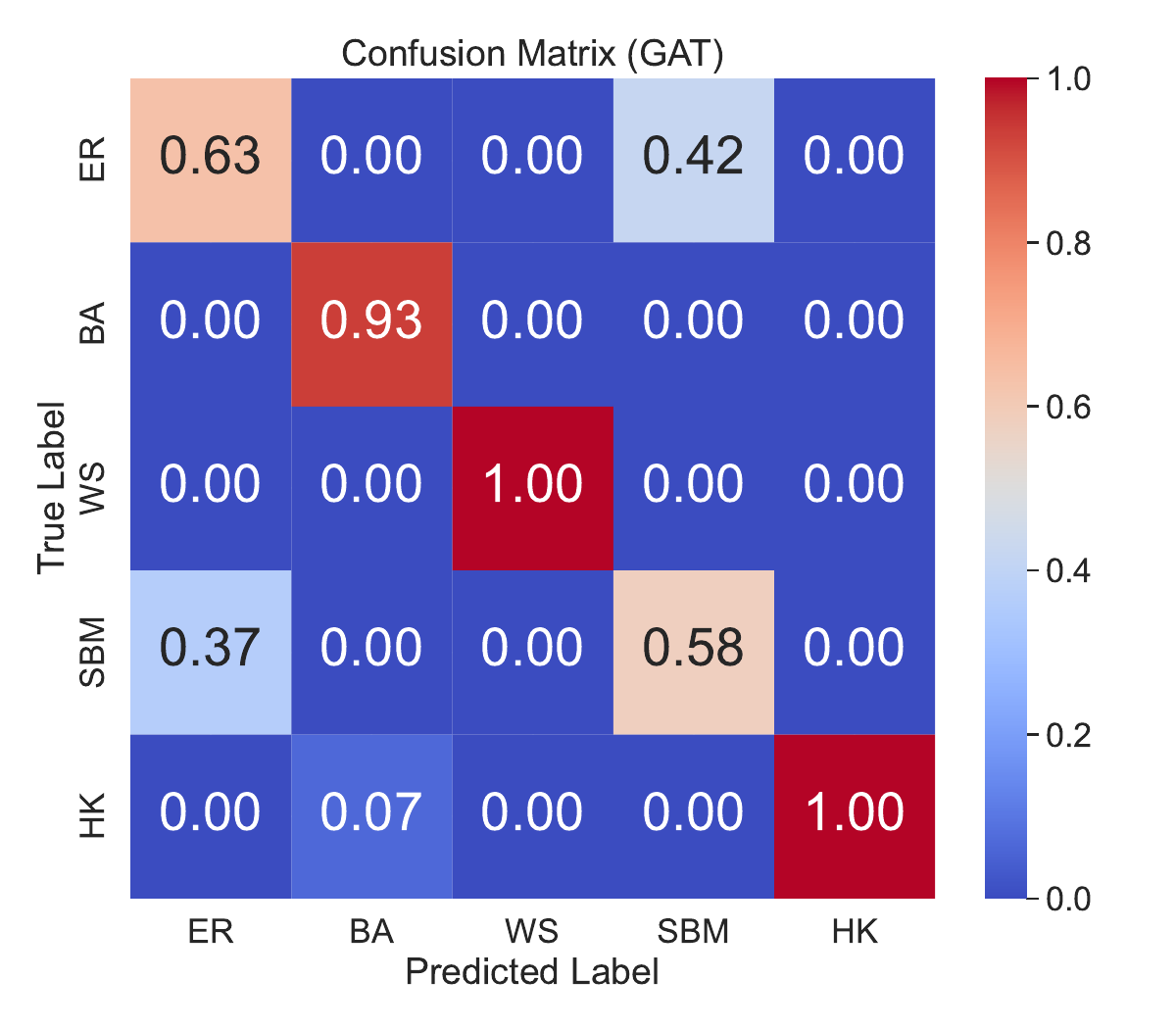}
    \caption{GAT}
\end{subfigure}
\caption{Confusion matrices for each GNN model. Strong performance is shown for most architectures, with the majority of classification errors occurring between Erdős–Rényi and Stochastic Block Models.}
\label{fig:confusion_matrix}
\end{figure}

\begin{figure}[H]
\centering

\begin{subfigure}[b]{0.475\textwidth}
    \includegraphics[width=\linewidth]{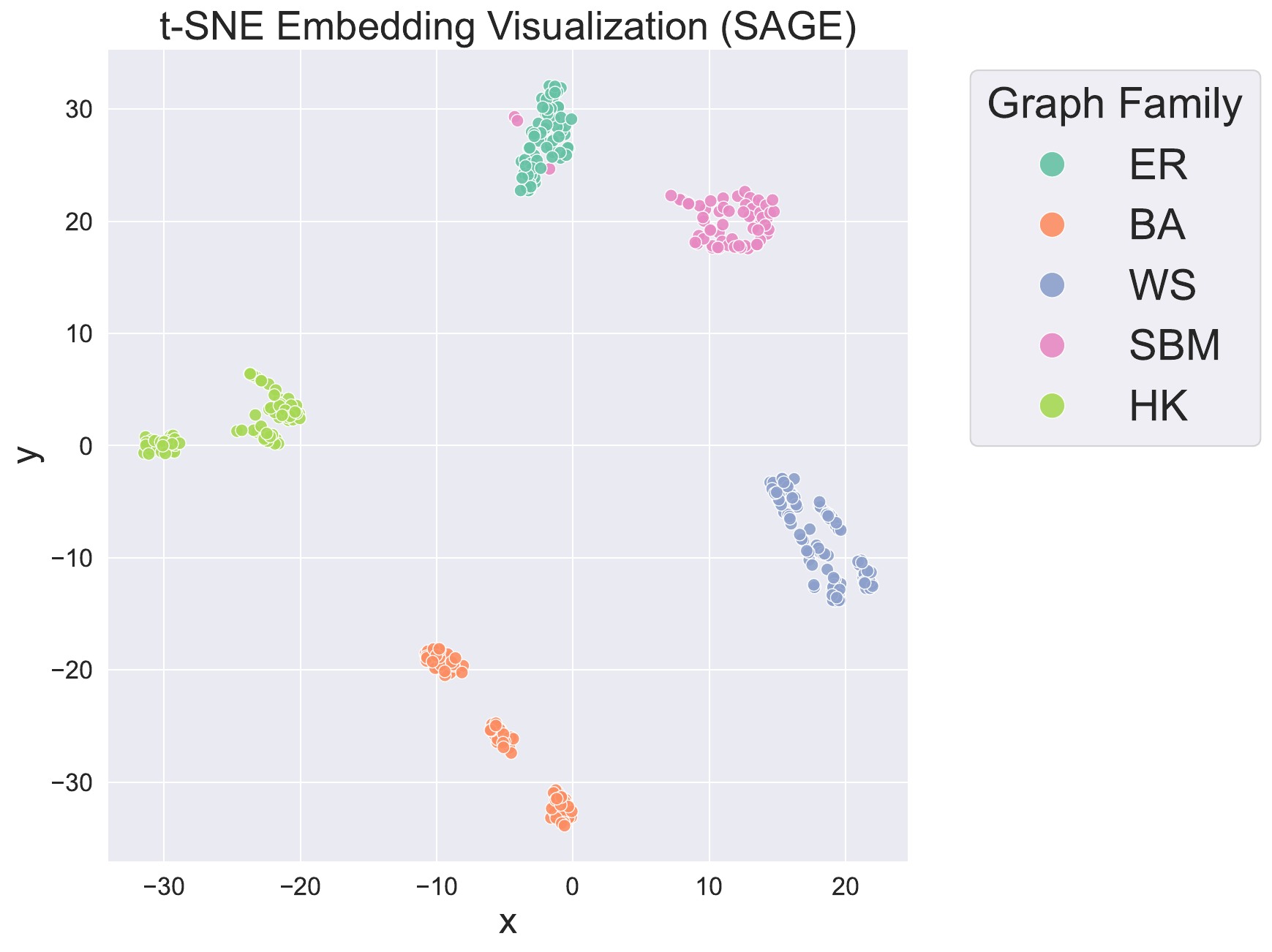}
    \caption{SAGE}
\end{subfigure}
\hfill
\begin{subfigure}[b]{0.475\textwidth}
    \includegraphics[width=\linewidth]{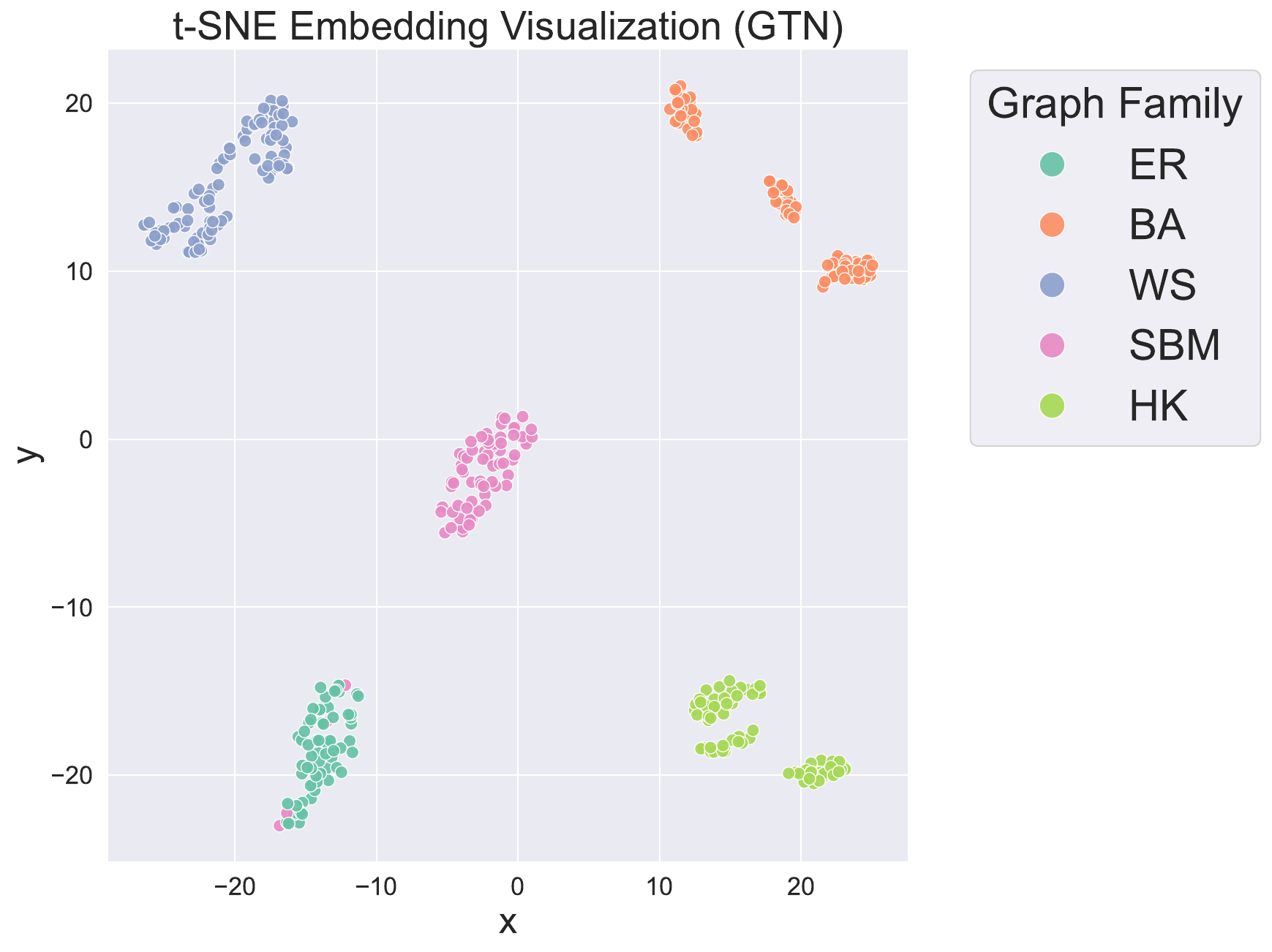}
    \caption{GTN}
\end{subfigure}

\vspace{0.25em}

\begin{subfigure}[b]{0.475\textwidth}
    \includegraphics[width=\linewidth]{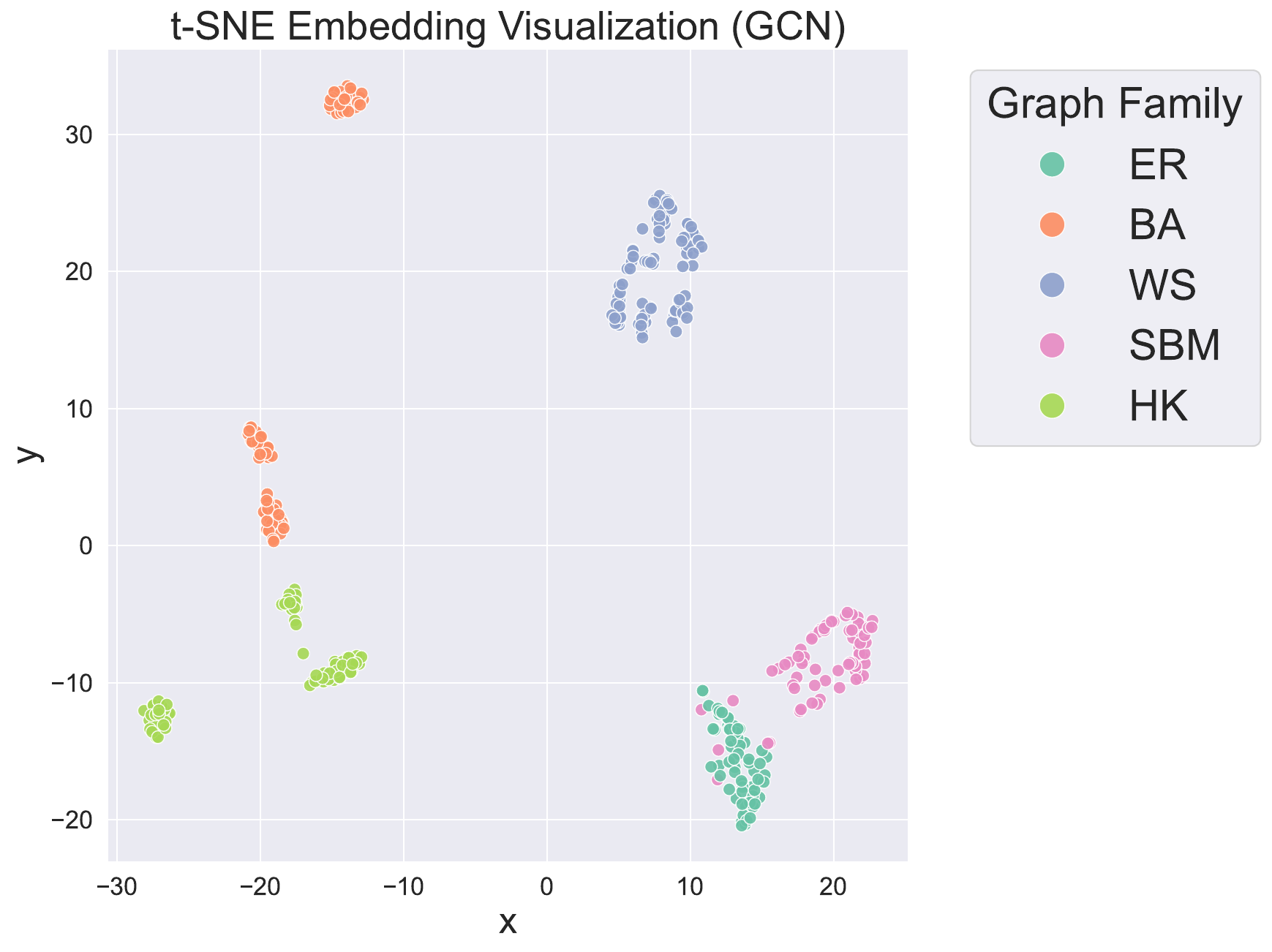}
    \caption{GCN}
\end{subfigure}
\hfill
\begin{subfigure}[b]{0.475\textwidth}
    \includegraphics[width=\linewidth]{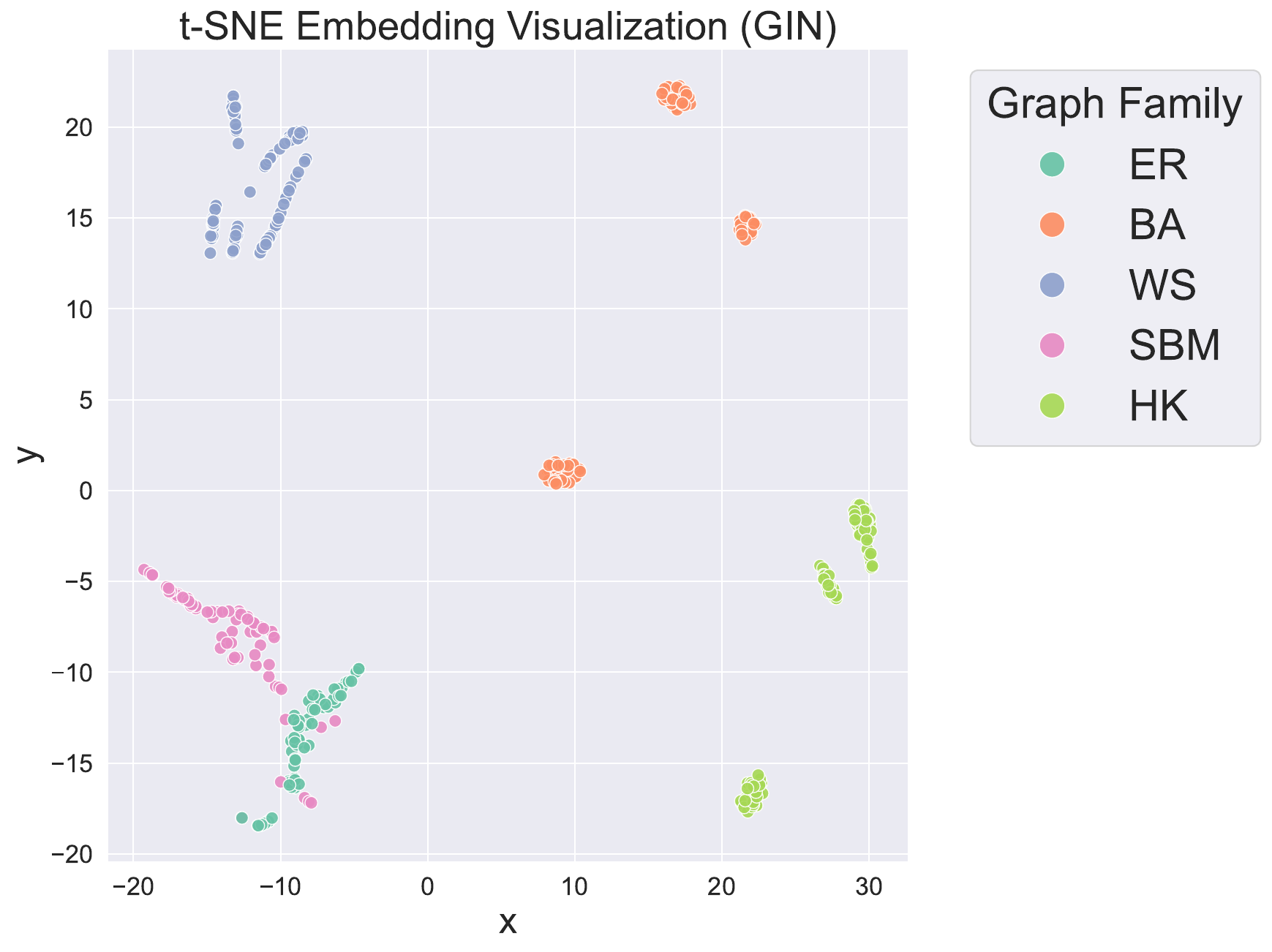}
    \caption{GIN}
\end{subfigure}

\vspace{0.25em}

\begin{subfigure}[b]{0.475\textwidth}
    \includegraphics[width=\linewidth]{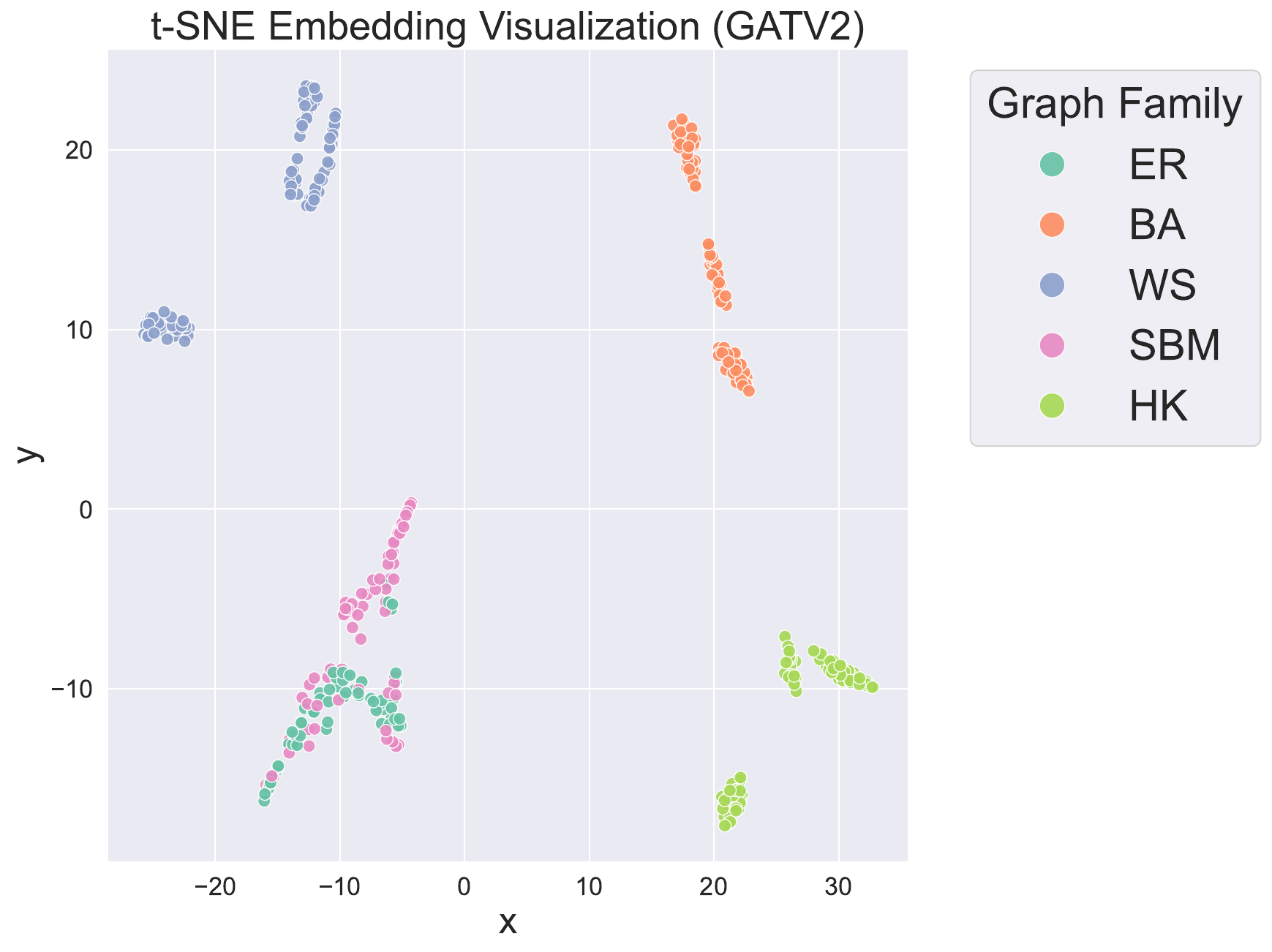}
    \caption{GATv2}
\end{subfigure}
\hfill
\begin{subfigure}[b]{0.475\textwidth}
    \includegraphics[width=\linewidth]{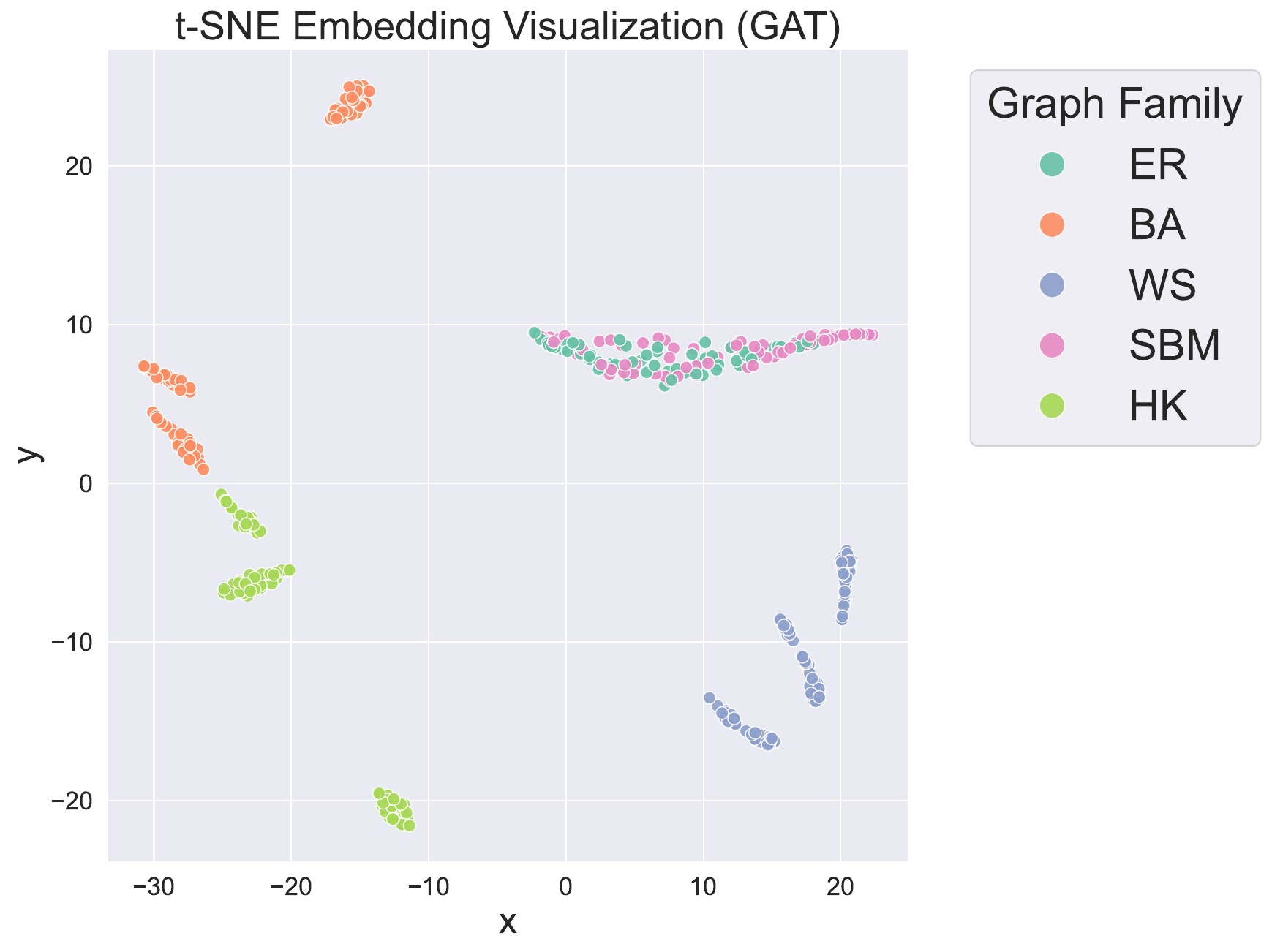}
    \caption{GAT}
\end{subfigure}

\caption{t-SNE plots of learned graph embeddings for each GNN model. Clear class separation is seen for most architectures, with visible overlap at the edges of classes known to be structurally similar.}
\label{fig:tsne}
\end{figure}

\begin{figure}[H]
\centering

\begin{subfigure}[b]{0.45\textwidth}
    \includegraphics[width=\linewidth]{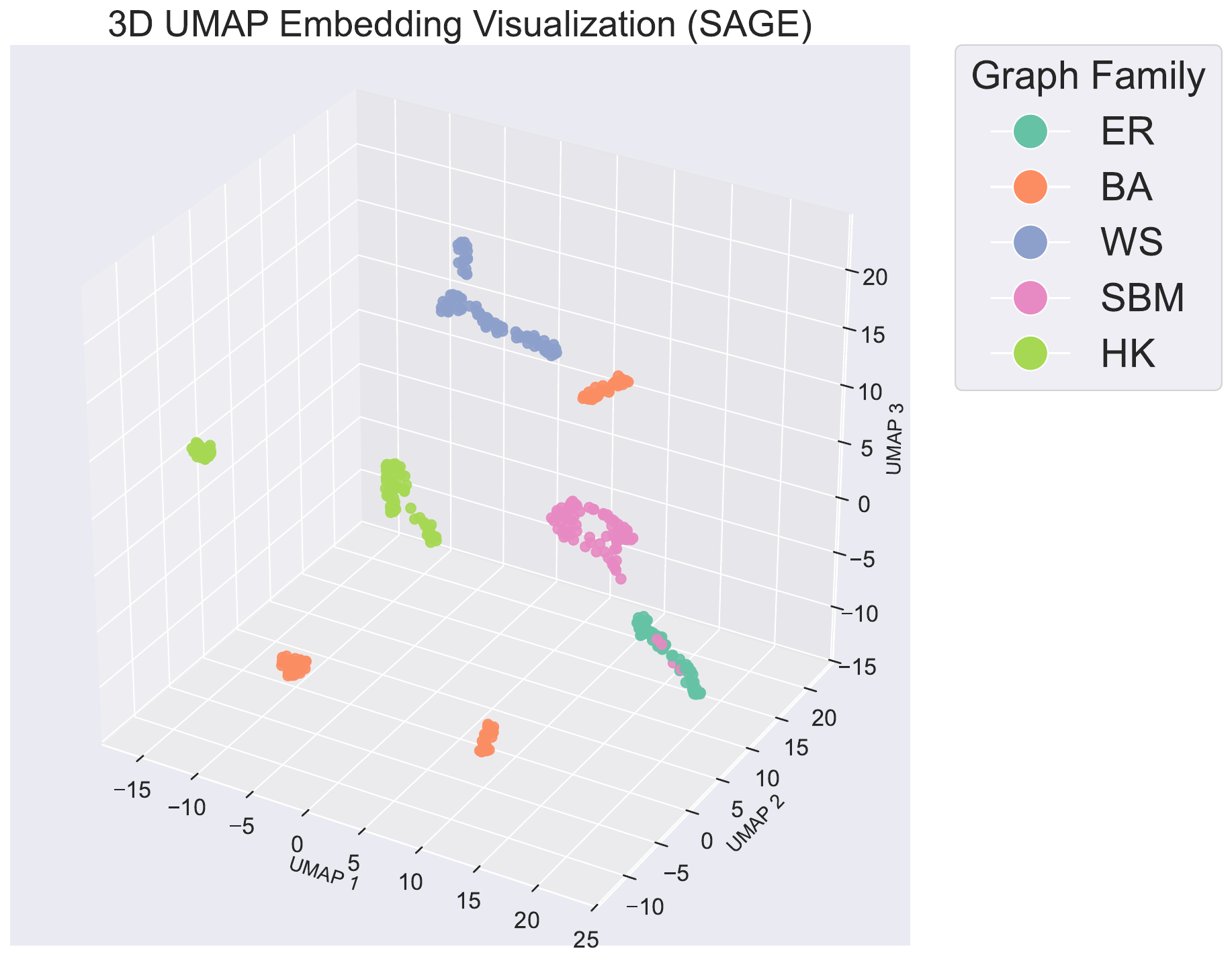}
    \caption{SAGE}
\end{subfigure}
\hfill
\begin{subfigure}[b]{0.45\textwidth}
    \includegraphics[width=\linewidth]{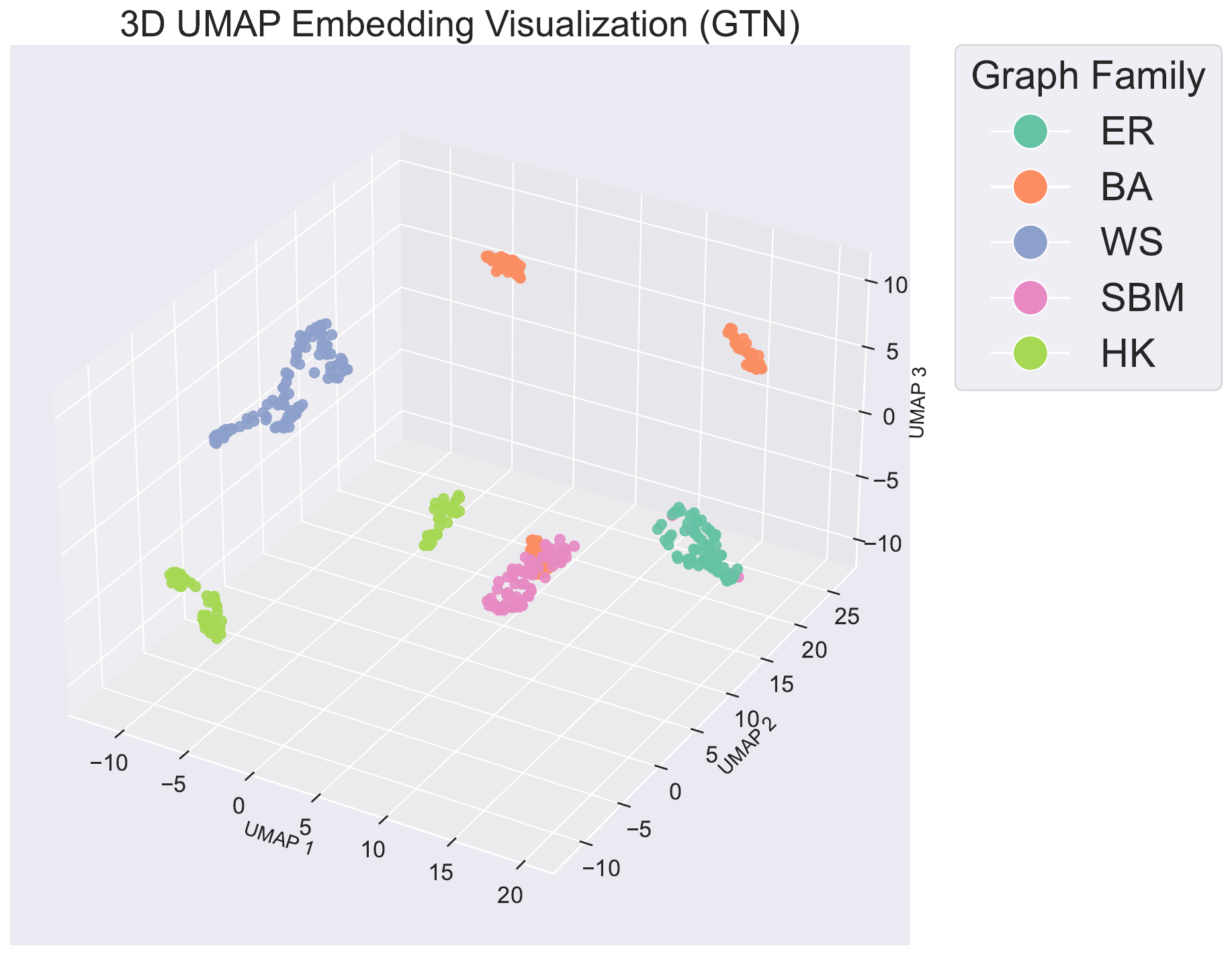}
    \caption{GTN}
\end{subfigure}

\vspace{0.25em}

\begin{subfigure}[b]{0.45\textwidth}
    \includegraphics[width=\linewidth]{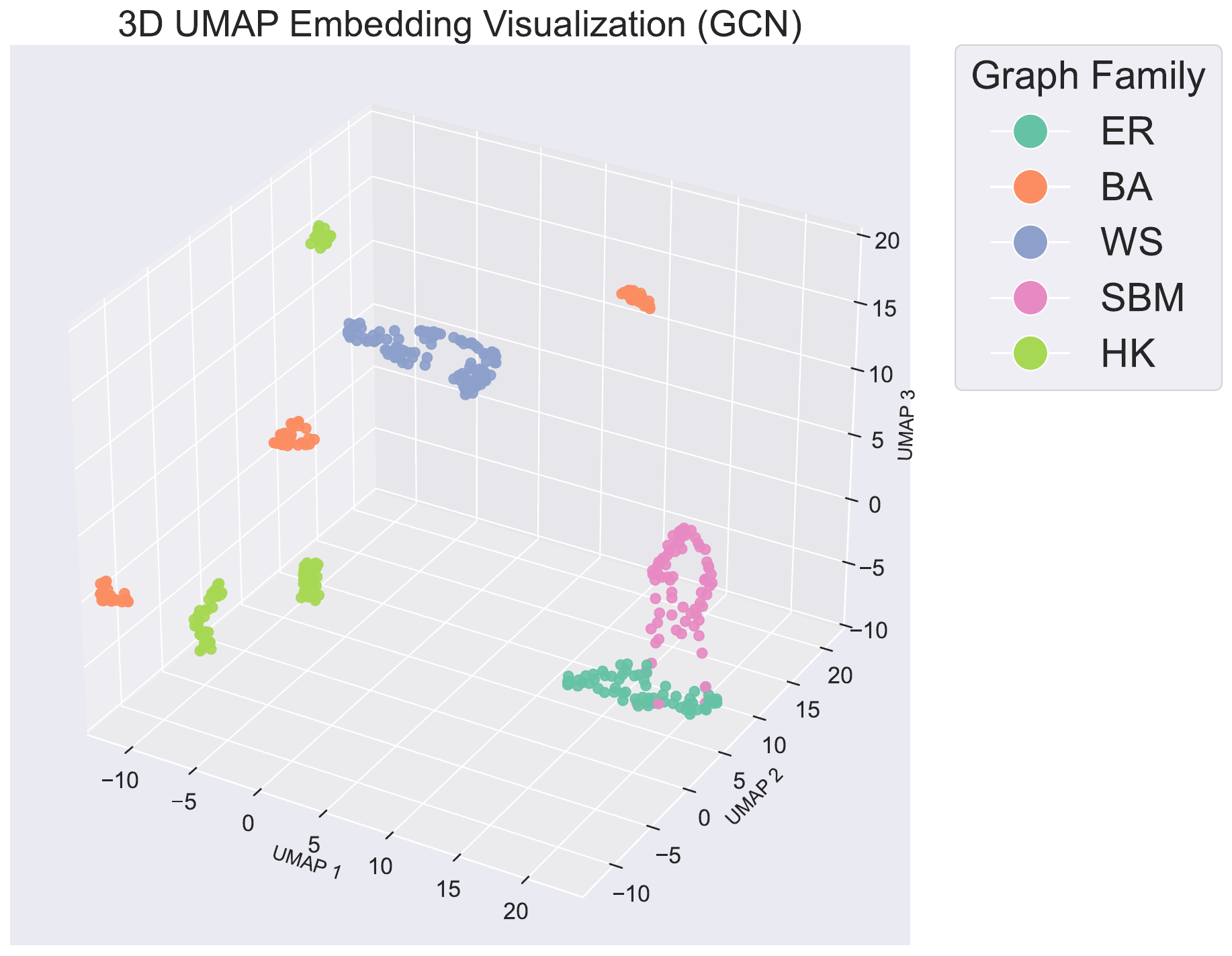}
    \caption{GCN}
\end{subfigure}
\hfill
\begin{subfigure}[b]{0.45\textwidth}
    \includegraphics[width=\linewidth]{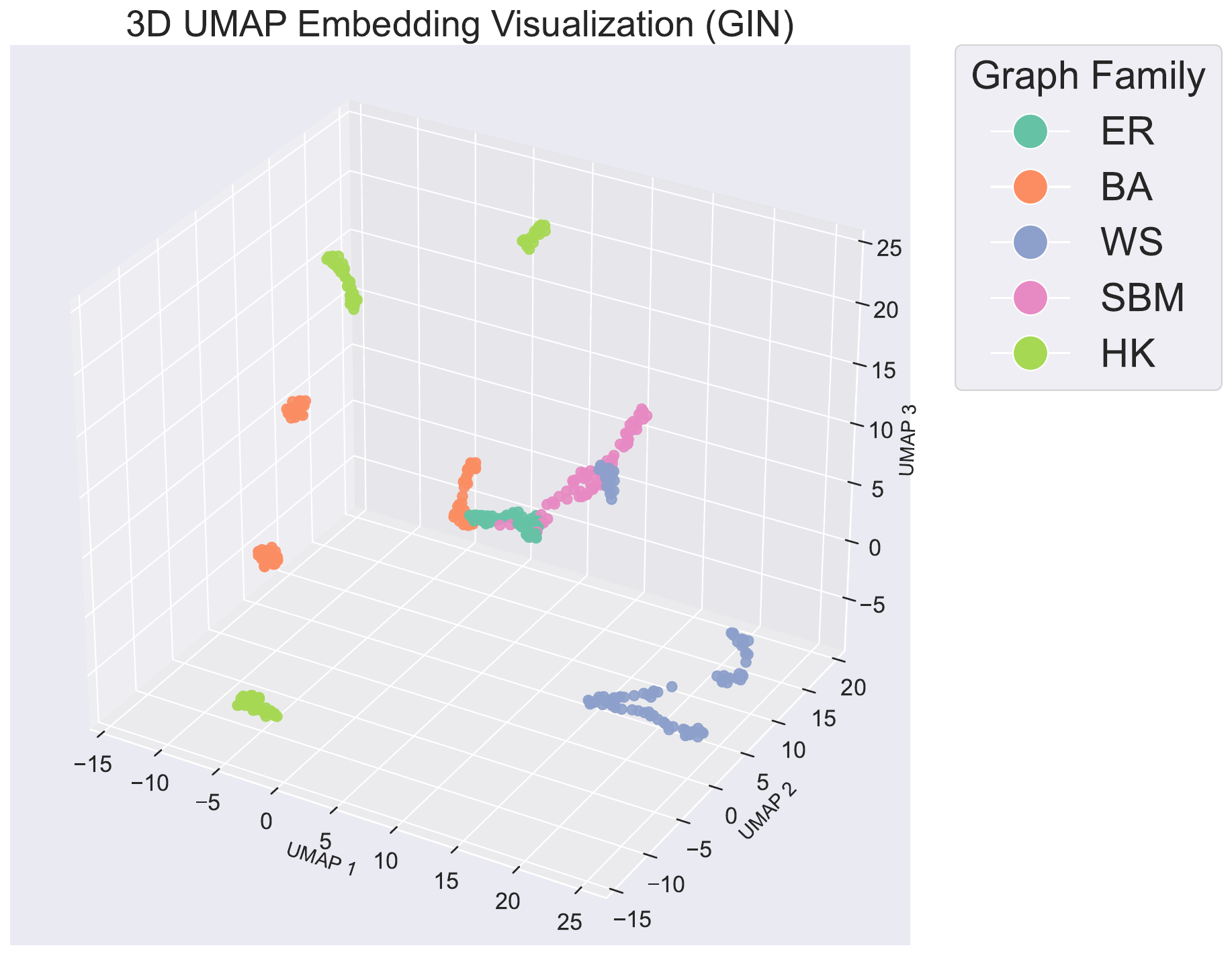}
    \caption{GIN}
\end{subfigure}

\vspace{0.25em}

\begin{subfigure}[b]{0.45\textwidth}
    \includegraphics[width=\linewidth]{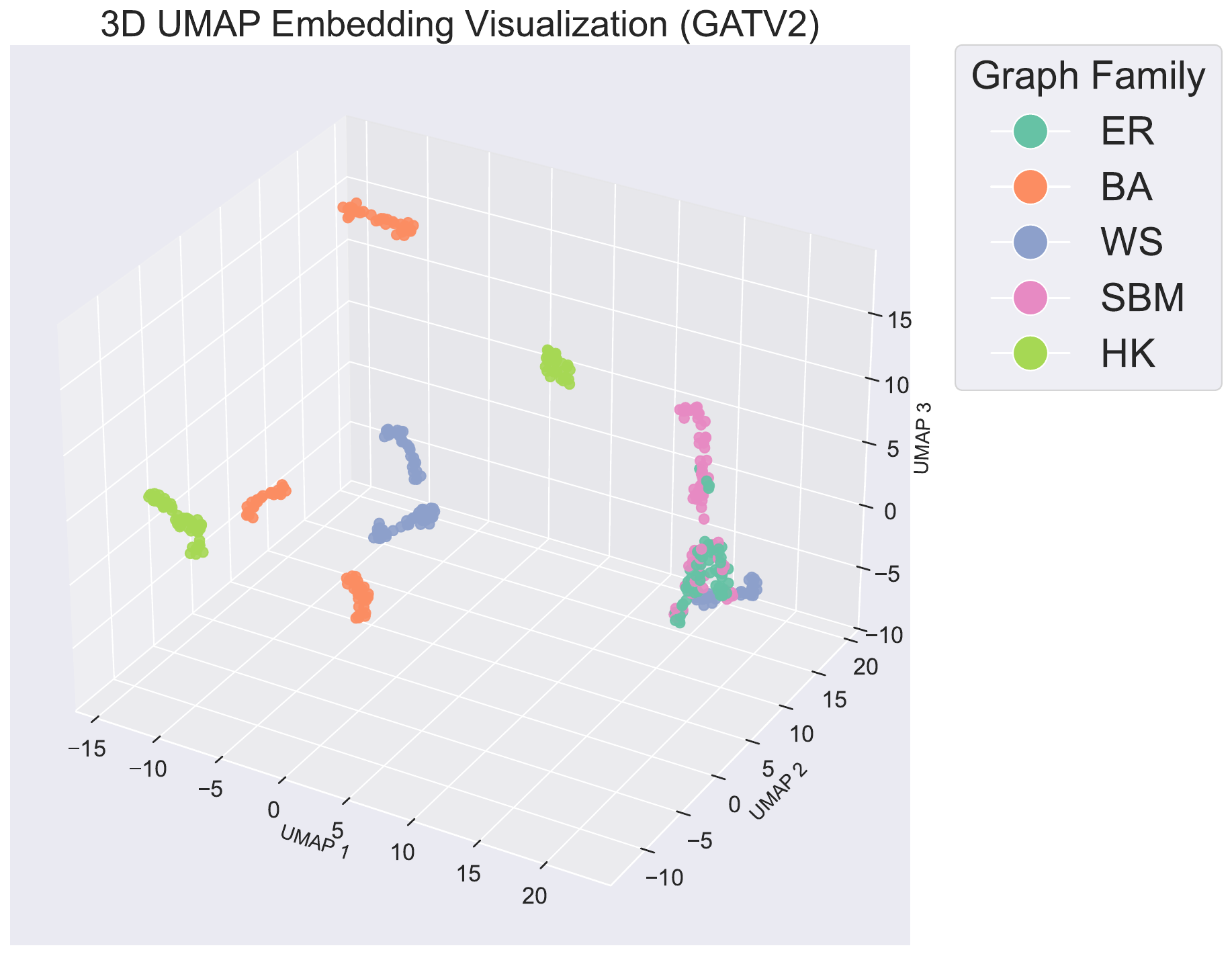}
    \caption{GATv2}
\end{subfigure}
\hfill
\begin{subfigure}[b]{0.45\textwidth}
    \includegraphics[width=\linewidth]{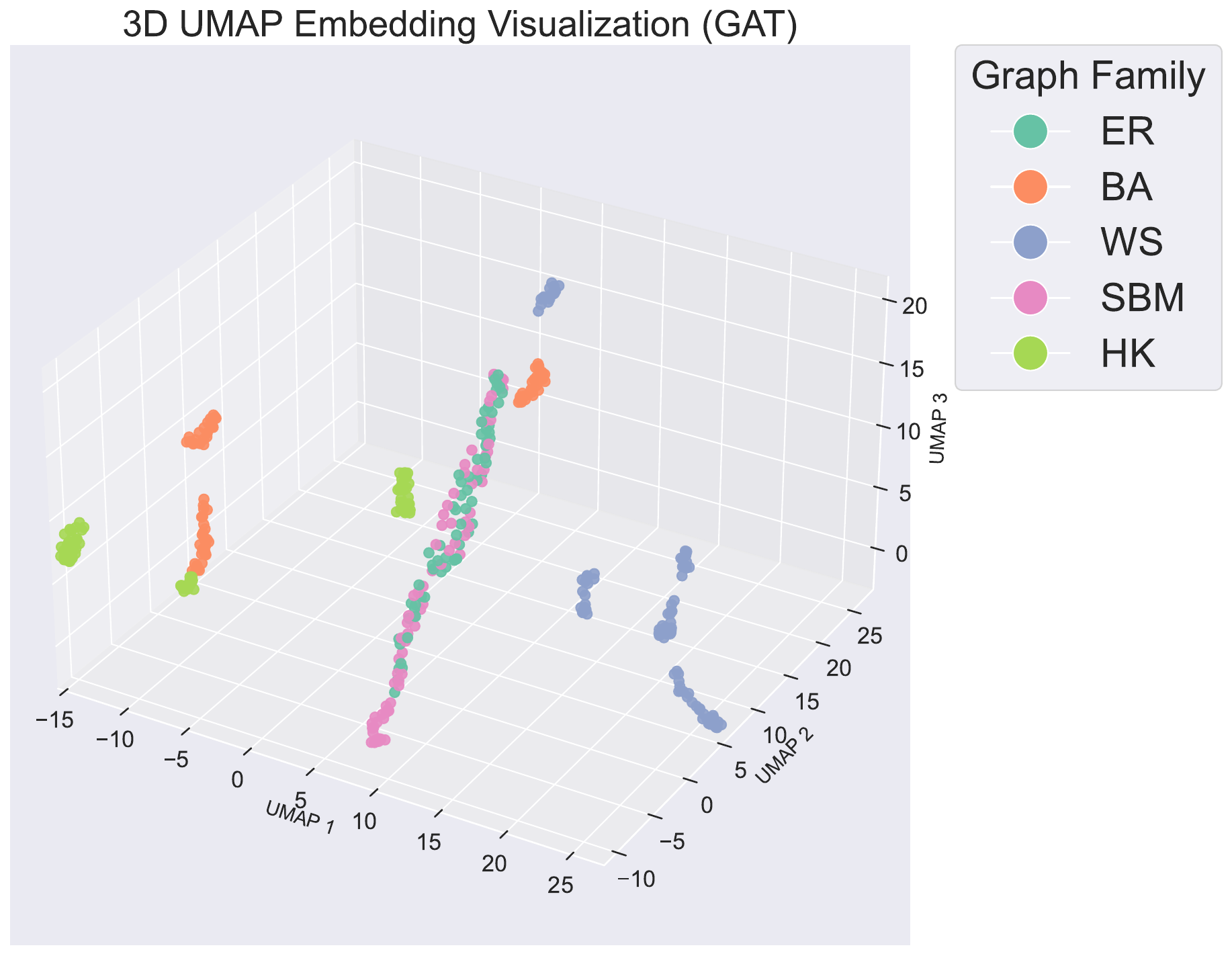}
    \caption{GAT}
\end{subfigure}

\caption{3-dimensional UMAP plots of learned graph embeddings for each GNN model. Clear class separation is seen for most architectures, with varying degrees of clustering tightness.}
\label{fig:umap}
\end{figure}

The most immediate observation from Table~\ref{tab:training_time} is the instability of the GAT model, which converged after just 7 epochs. This limited training horizon likely contributed to its poor performance. The rest of Table \ref{tab:training_time} evidences the trade-offs between discriminative power and training time. Although GTN and SAGE achieved virtually identical results in terms of performance metrics, the SAGE model achieved its best results in two-thirds of the epochs, and one-third of the overall time. This highlights a significant advantage of the inductive reasoning and sample-based aggregation carried out by SAGE, making it a strong choice for such a task. In summary, GraphSAGE offers the best balance of performance, efficiency, and training stability, while GTN achieves state-of-the-art performance at the cost of increased training time. Models with dynamic or global aggregation mechanisms consistently outperform those relying on static or local updates. To support interpretability of results and highlight differences in model behaviours, visual summaries of each model's performance are provided. These include confusion matrices (Figure~\ref{fig:confusion_matrix}) and t-SNE/UMAP plots (Figures~\ref{fig:tsne} and \ref{fig:umap}) of the learned embeddings. These plots illustrate clustering behaviours and class-level separability, with UMAP providing a three-dimensional perspective.

Across all visualisations, the top-performing models demonstrated very strong class separation. The GraphSAGE, GTN, GCN, and GIN models each showed perfect or near-perfect results in the confusion matrices, with overall performance separated by rates of predicting Erdős–Rényi graphs as Stochastic Block Models. The underperforming GAT and GATv2 models demonstrated evidence of bidirectional misclassification between these two families, as well as additional misclassifications of Barabási-Albert graphs as Holme-Kim.

The embeddings plots provided further evidence of the abilities of the top-performing models to provide class separation. In particular, the t-SNE plots obtained from the GraphSAGE and GTN models demonstrate excellent clustering, with only slight evidence of overlap. This is contrasted with weaker performers such as GAT, where near complete overlap between Erdős–Rényi and Stochastic Block Model is evident. The UMAP output provides an enhanced sense of the global structure of the embeddings. While tSNE does provide accurate grouping, and a strong sense of local clustering patterns, the differences between these clusters is better illustrated by UMAP.

\section{Discussion}\label{disc}
This study demonstrates the importance of architecture design in GNN-based approaches to the effective classification of complex graphs. The results indicate that models which incorporate inductive learning or global attention mechanisms present consistent excellent performance across both standard performance metrics and embedding-based visualisations. 

GraphSAGE in particular stands out, not only for its accuracy, but also for its strong computational efficiency and stability. The use of learning through sampled neighbourhoods makes it well suited to the diverse graph structures featured in the experimental dataset. GTN, while achieving strong comparable performance, incurs a higher computational cost due to the reliance upon global attention networks. This highlights a key trade-off between complexity and efficiency, and demonstrates the scalability of GraphSAGE.

GIN and GCN models also exhibited strong performances, although their limitations were evidenced through their drop in performance when classifying Erdős–Rényi and Stochastic Block Models. In the case of GIN, it suggests that the sum-based aggregation approach, although theoretically sound, fails to fully capture the modular Block Model structure. The difficulties exhibited by GCN in this area suggests that the static averaging used in neighbourhood aggregation dilutes discriminative features of local community structures. 

GAT and GATv2 underperformed relative to the other GNN architectures. These models rely on local attention mechanisms which, while highly effective in node-level classification tasks, may be insufficient for determining global structures in structurally similar and complex graphs, where long range dependencies play a significant role in class separation. Although GATv2 showed improvements over GAT, it still failed to fully separate families with overlapping local properties.

The SVM model’s relatively poor performance reinforces the importance of message passing and structural learning for this task. However, it also highlighted the continued value of handcrafted features when appropriately selected and pruned. It's non-trivial classification performance makes it a useful point of comparison against the specialised GNN architectures.

For all architectures, the Erdős-Rényi family proved the most difficult to classify, especially for the GAT-based models. Misclassifications typically involved confusion with Stochastic Block Models, likely due to parameter configurations that yielded visually and structurally similar graphs.

In contrast, architectures seemed to have very little issue in discriminating between Barabási-Albert and Holme-Kim graphs, despite their strong structural similarities. This suggests that the local clustering exhibited by Holme-Kim may have contributed towards models using local structures to make strong predictions about global attributes.

Trends in classification confusion were reflected in the t-SNE and UMAP plots, with overlap in clusters existing as expected between graphs known to be structurally similar. This alignment provides good evidence of strong structural insights obtained from the top-performing models.

In summary, the experimental results of this study suggest that message passing frameworks which balance both local and global structures provide superior performance. Both the inductive sampling approach of GraphSAGE and the global attention approach of GTN offering excellent results, offering a good balance between performance, robustness and efficient. Overall, the practical benefits offered by GraphSAGE, including faster training and lower computational costs, make GraphSAGE particularly well suited for scalable applications in this area.

This study provides a strong case for the use of synthetic graph datasets
for benchmarking GNN-based architectures. Through the modular pipeline provided,
it allows for strong insight into the strengths and weaknesses of each architecture,
allowing for appropriate scaling. The graphs featured in the dataset represent a diverse range of node and edge sizes, demonstrating good generalisability of approach.

This synthetic nature of the dataset does serve as a limitation to the study. While an inherently clean and noise-free dataset is very useful for demonstrating
the full capacities of the architectures featured, the lack of noise as found in real- world datasets means that performance may not necessarily translate directly between dataset formats. However, using a synthetic dataset as a baseline for performance and explainability allows for architectures to be benchmarked in a controlled environment.

The ability to manually dictate network structures using probability-based generative parameters allowed for datasets to be developed in line with requirements in terms of structural similarities.

\section{Conclusion}\label{conc} 
This study provides a comprehensive evaluation of six GNN architectures for graph classification using a novel synthetic benchmark dataset, constructed from five structurally similar generative families. A hybrid feature approach was used, combining handcrafted graph-theoretic features with GNN-based message passing. A baseline SVM model trained solely on handcrafted features was also included for comparison.

In this study, the GraphSAGE and GTN models consistently outperformed the other architectures, providing strong evidence in favour of the inductive reasoning and global attention field approaches employed by the models for this task. GraphSAGE in particular stood out due to its exceptional performance in the context of its relatively small size and short training time. GIN and GCN models also showed strong results, with efficient trade-offs between model performance and complexity. GAT-based models significantly underperformed, with difficulties in effectively discriminating between structurally similar graph families.

The performance of the GNN-based models, compared to baseline, highlight the effectiveness of the message-passing operation. The training times and model sizes relative to performance demonstrated the need for architectural choices to balance expressiveness, efficiency, and stability. Although architectures were relatively shallow, the performance metrics of the top performers illustrated that the depth used was sufficient to achieve effective and consistent discrimination. 

The use of lightweight features for integration limits their discriminative power in such complex graphs, as evidenced by the SVM performance. This is somewhat dictated by the scale of the search space, which makes more complex features computationally prohibitive to calculate. The flexibility and scalability of the pipeline as presented allows for more complex features to be integrated, allowing for the balance between expressiveness and computational cost to be tailored to the use case.

Future work could extend this approach to include larger or real-world graphs. Investigating the role of more computationally expensive features (e.g., spectral or topological descriptors) could improve discrimination at the cost of complexity. Additionally, architecture-specific hyperparameter search spaces may further enhance optimisation outcomes.

\section*{Declarations}

\subsection{Code and Data Availability}

The full source code and dataset created in this study are publicly available at: \\

\texttt{\href{https://github.com/j-dyer-code/Synthetic-Graph-Classification-GNN}{https://github.com/j-dyer-code/Synthetic-Graph-Classification-GNN}}

Please do not hesitate to contact the authors if you have any questions.
%%===========================================================================================%%
%% If you are submitting to one of the Nature Portfolio journals, using the eJP submission   %%
%% system, please include the references within the manuscript file itself. You may do this  %%
%% by copying the reference list from your .bbl file, paste it into the main manuscript .tex %%
%% file, and delete the associated \verb+\bibliography+ commands.                            %%
%%===========================================================================================%%
\bibliographystyle{ACM-Reference-Format}
\bibliography{sn-bibliography}
%\bibliography{sn-bibliography}% common bib file
%% if required, the content of .bbl file can be included here once bbl is generated
%%\input sn-article.bbl

\end{document}